\documentclass{article}

\PassOptionsToPackage{numbers, compress}{natbib}

\usepackage[preprint]{neurips_2026}

\usepackage[utf8]{inputenc} 
\usepackage[T1]{fontenc}    
\usepackage{hyperref}       
\usepackage{url}            
\usepackage{booktabs}       
\usepackage{amsfonts}       
\usepackage{amsmath}        
\usepackage{nicefrac}       
\usepackage{microtype}      
\usepackage{xcolor}         
\usepackage{graphicx}       
\usepackage{xspace}         
\usepackage{multirow}       
\usepackage{longtable}      
\usepackage{array}          
\usepackage{afterpage}      
\usepackage{wrapfig}        
\usepackage{placeins}       

\definecolor{semanticblue}{RGB}{139, 187, 219}
\definecolor{motionorange}{RGB}{243, 164, 97}

\newcommand{\ourmodel}{AFUN\xspace}
\newcommand{\ourdataset}{\ourmodel dataset\xspace}
\newcommand{\ourmodeltestset}{\ourmodel test set\xspace}
\newcommand{\curverep}{Bézier spline curve\xspace}

\newcommand{\loss}{\mathcal{L}}
\newcommand{\lossSam}{\loss_{\mathrm{sam3}}}
\newcommand{\lossCurve}{\loss_{\mathrm{curve}}}

\newcommand{\wSam}{\lambda_{\mathrm{sam3}}}
\newcommand{\wCurve}{\lambda_{\mathrm{curve}}}

\newcommand{\matchset}{\mathcal{M}}

\newcommand{\vB}{\mathbf{B}}
\newcommand{\vP}{\mathbf{P}}

\newcommand{\pred}[1]{\widehat{#1}}
\newcommand{\gt}[1]{#1^{\star}}

\newcommand{\mqsem}[1]{\langle \mathrm{mq}^{s}_{#1} \rangle}
\newcommand{\mqmot}[1]{\langle \mathrm{mq}^{m}_{#1} \rangle}
\newcommand{\mqsemset}{\mathbf{mq}^{s}}
\newcommand{\mqmotset}{\mathbf{mq}^{m}}

\graphicspath{{figures/}}

\newcommand{\authorskip}{\hspace{3mm}}
\newcommand{\institutionskip}{\hspace{2mm}}

\title{\ourmodel: Towards an Affordance Foundation Model for Functionality Understanding}

\author{
  Zhaoning Wang\textsuperscript{1 *} \authorskip
  Yi Zhong\textsuperscript{1 *} \authorskip
  Jiawei Fu\textsuperscript{2} \authorskip
  Henrik I. Christensen\textsuperscript{2} \authorskip
  Jun Gao\textsuperscript{1, 3}
  \vspace{1.5mm} \\
  \small{
  \textsuperscript{1}University of Michigan \institutionskip
  \textsuperscript{2}University of California, San Diego \institutionskip
  \textsuperscript{3}NVIDIA}
}

\begin{document}

\maketitle

\let\thefootnote\relax
\footnotetext{
\small
\textsuperscript{*} Equal contribution.
}

\begin{abstract}
  \vspace{-2mm}
Affordance understanding bridges visual perception and physical action, serving as an explainable interface for robot manipulation in open and unstructured real-world environments.
Yet, building an affordance foundation model that not only understands \emph{where} and \emph{how} the interaction should happen, but also generalizes across diverse environments, objects, and tasks, remains a long-standing research challenge.
Existing methods typically address only part of this challenge, either localizing task-relevant regions without specifying executable motion, or predicting motion but with limited scalability.
In this paper, we present \textbf{\ourmodel}, a step towards an affordance foundation model for functionality understanding. From a single RGB-D observation and a language task description, \ourmodel predicts a task-conditional functional mask (\emph{where} to interact) and a 3D post-contact motion curve (\emph{how} to interact).
To support open-world generalization, we build a large-scale standardized data pipeline that converts heterogeneous robot, human, simulation, and real-world scan data into a shared affordance schema with language, masks, and object-centric 3D motion labels.
We evaluate \ourmodel from three aspects: for affordance segmentation, \ourmodel outperforms all baselines by a large margin across 8 test sets from 4 benchmarks, improving mean gIoU/cIoU by \textbf{+23.9/+26.3}; for contact-point prediction, it predicts substantially more accurate points, with a \textbf{12.7--61.3\%} hit-rate gain over the best baseline; and for 3D motion, it achieves the best performance on all three test sets.
\ourmodel can be deployed for real-world robot manipulation without finetuning for robot embodiment or using task-specific heuristics, demonstrating the ability to adapt to open-world affordance tasks. Project page: \url{https://www.zhaoningwang.com/AFUN}

\end{abstract}

\vspace{-2mm}
\section{Introduction}
\vspace{-1mm}
\label{sec:introduction}

Imagine stepping into a brand-new bedroom; a human can immediately understand \emph{which} object can do \emph{what}, and \emph{how} to do it. For instance, a drawer can be opened or closed from its handle, and a human can identify the exact grasping location before the actual action. 
This concept of visual \emph{affordance}~\cite{gibson1979ecological} to understand objects' functionalities underpins human's capability to perform daily tasks in unstructured real-world environments~\cite{affordancer1,tang2025uad}. 
In robotics and embodied AI, affordance understanding serves as a crucial and explainable interface between visual understanding and physical action.
Yet, building a foundation model for affordance understanding that can scale across diverse environments, objects, and tasks is a long-standing research challenge.

There are three interconnected requirements when building such an affordance foundation model.
\textbf{(I)} The dataset used to train the model must reflect the diversity of real-world manipulation tasks to enable generalization, rather than being collected from narrow domains or a closed set of object categories.
\textbf{(II)} The model needs to accurately produce instruction-conditioned segmentation masks: not only locating where robots can interact with, but also adapting to the instruction, since the same object affords different regions under different tasks. 
\textbf{(III)} To make the interaction actionable for robots, the model must further predict \emph{how} the interaction should be performed, with a 3D motion representation that a robot can follow. The 3D motion should remain expressive enough to capture diverse behaviors and structured enough for stable supervision and robot execution.

In practice, however, existing affordance methods focus mainly on the second requirement alone, formulating the problem as static segmentation~\citep{ragnet,gloverpp}, keypoint detection~\citep{yuan2024robopoint}, or reasoning-based grounding~\citep{affordancer1}.
These approaches can localize interaction regions, but they do not characterize how the object should move after interaction.
For the methods focusing on the third requirement, some predict motion in 2D~\citep{xu2025a0,bahl2023vrb}, leaving robot execution ambiguous when lifting into 3D, while others~\citep{yuan2024general} require heuristic localization of actionable objects.
Beyond limitations on each modality, most current deep-learning models for affordance understanding~\citep{xu2025a0,bahl2023vrb,chen2025vidbot,yuan2024general} still fall short of open-world generalization due to small-scale datasets with limited diversity.

\begin{figure*}[t]
  \centering
\includegraphics[width=\linewidth]{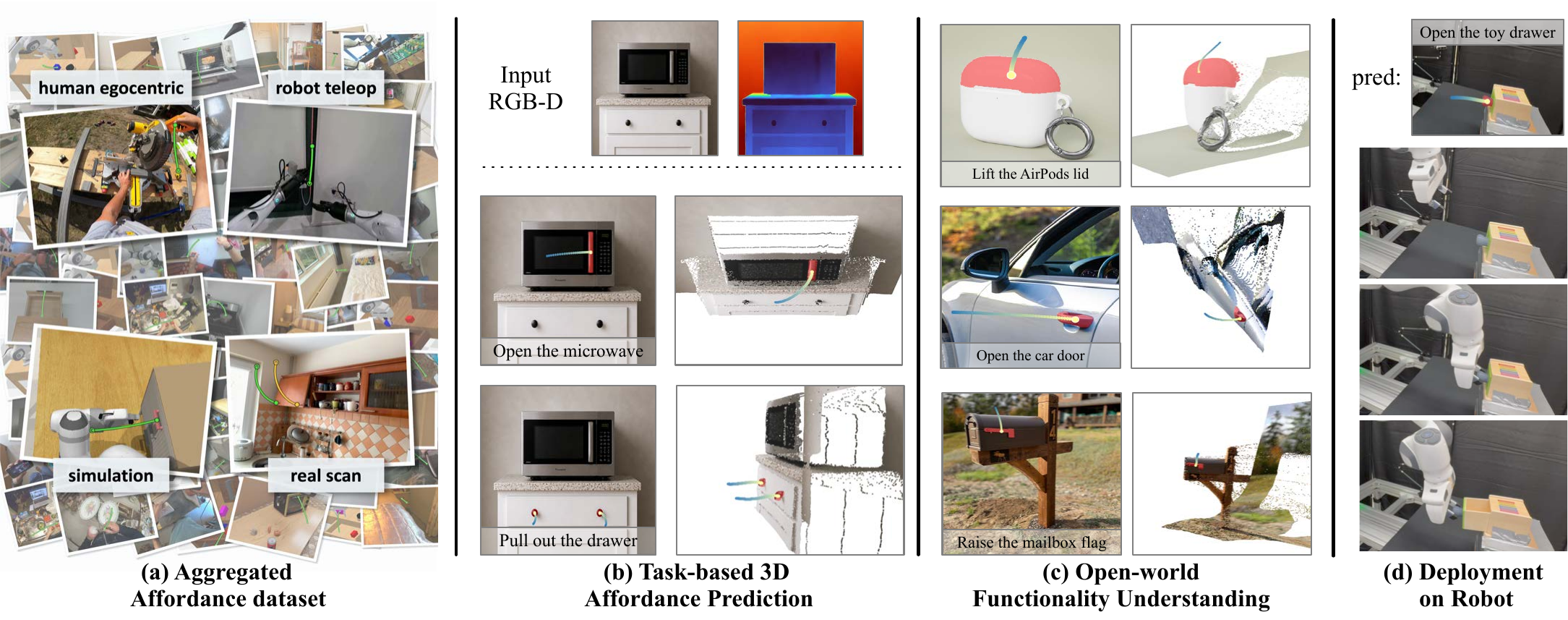}
  \caption{
    \textbf{Overview of \ourmodel.}
    \textbf{(a)} We first build a data pipeline to gather a large-scale diverse dataset for affordance understanding.
    \textbf{(b)} With such a dataset, we then train \ourmodel to predict a task-conditional functional segmentation mask and a 3D motion trajectory, conditioned on an RGB-D observation and a language task phrase.
    \textbf{(c)} \ourmodel can generalize to open-world images for functionality understanding and \textbf{(d)} is directly deployable to the real robot for manipulation.
  }
  \label{fig:teaser}
\end{figure*}

To address these gaps, we present \textbf{\ourmodel}, a step toward an open-world affordance foundation model.
First, we build a large-scale standardized data pipeline that converts public robot, human, and simulation datasets into coherent affordance data with task descriptions, functional masks and 3D motions, extending the current affordance dataset towards one of the largest public affordance datasets to date (Figure~\ref{fig:teaser} (a)). 
Then, we introduce a unified affordance foundation model that jointly predicts \emph{where} to act (functional segmentation) and \emph{how} the interaction should happen (3D motion, represented as a \curverep), conditioned on text instructions from users and robot-native RGB-D observations, as shown in Figure~\ref{fig:teaser} (b).
The mask can then be unprojected into 3D points for robots to perform action leveraging downstream grasping modules such as AnyGrasp~\citep{fang2023anygrasp}.

We evaluate \ourmodel on 8 segmentation-based affordance benchmarks and 3 motion-based benchmarks.
\ourmodel reaches \textbf{69.3} mean segmentation gIoU, compared with \textbf{45.4} for the strongest segmentation baseline~\citep{affordancer1}; for motion prediction, it surpasses standalone baselines~\citep{xu2025a0,bahl2023vrb,chen2025vidbot} by a substantial margin.
\ourmodel also demonstrates strong generalization capability when qualitatively evaluated on open-world images (Fig.~\ref{fig:teaser}(c)).
Furthermore, we deploy \ourmodel on a real robot for manipulation. Without any robot-specific finetuning, \ourmodel can predict precise mask and motion for robot to plan and execute a successful path for manipulation, as illustrated in Fig.~\ref{fig:teaser} (d) and Fig.~\ref{fig:robotic_demo}.

\vspace{-2mm}
\section{Related Work}

\label{sec:related_work}

\vspace{-2mm}
\paragraph{Affordance localization.}

Affordance localization mainly asks \emph{where} a task-specified interaction is possible with various grounding representation.
Dense 2D methods cover classical instance- and part-level segmentation~\citep{nguyen2017object_crf,do2018affordancenet}, weakly supervised cross-view grounding~\citep{luo2022learning,li2023locate,jang2024intra,xu2025partprior}, egocentric- and human-video mask supervision~\citep{li2024precise,heidinger2025_2handed,gloverpp}, LISA-style SAM grounding from LLM hidden states~\citep{lai2023lisa,qian2024affordancellm}, two-stage VLM-to-segmenter pipelines using LLM-emitted coordinates or boxes~\citep{affordancer1,li2024manipllm,huang2024manipvqa,chen2024worldafford}, and language-conditioned SAM-style benchmarks and decoders~\citep{ragnet,instructpart,jiang2025affordancesam,huang2025bitalign}.
Sparse alternatives predict contact points or keypoints, including language-conditioned image keypoints~\citep{yuan2024robopoint}, 3D keypoint quadruplets encoding contact location and direction~\citep{fakp2025}, and cross-category contact transfer by semantic correspondence or retrieval~\citep{ju2024robo_abc,kuang2024ram}.
3D approaches use point clouds, from foundational benchmarks and 2D-to-3D interaction grounding~\citep{deng20213daffordnet,yang2023iagnet} to open-vocabulary and language-conditioned 3D-MLLM grounding~\citep{nguyen2023openad,shao2025great,lu2025geal,zhu2025lmaffordance3d,chu2025_3daffordllm,yu2025seqafford,affordbot}, cross-instance or object-to-object transfer~\citep{o3afford,semantic_funcmap}, and video-driven MLLM learning from human-object interaction~\citep{videoafford,vagnet}.
Hand-centric work localizes functional grasps and dexterous contacts, from category-level grasp generation~\citep{corona2020ganhand} and language-guided task-oriented grasping~\citep{affordgrasp,mllm_afford_grasp,yu2024uniaff} to dexterous and finger-specific affordance prediction~\citep{afford_dex_grasp,fsag}.
These lines leave post-contact motion---\emph{how} the object should move---largely unspecified, motivating \ourmodel's joint mask-and-motion formulation.

\vspace{-3mm}
\paragraph{Motion representations for affordance.}
Motion-focused affordance work asks \emph{how} the object should move after contact, with representations varying in granularity and structure.
Some methods use discrete prompts or parametric articulation: elementary push/pull types tied to actionable regions~\citep{mo2021where2act}, scene-level functional categories with motion type and axis~\citep{delitzas2024scenefun3d}, or openable-part motion-parameter regression for articulated objects~\citep{opd,opdmulti,mopd}.
Others predict continuous interaction geometry, including dense visual action trajectories and per-point 3D articulation flow~\citep{wu2022vatmart,flowbot3d,flowbotpp,toolflownet}, egocentric contact heatmaps and 6-DoF object trajectories~\citep{yoshida2024text,yoshida2025_6dof}, hand and wrist trajectories distilled from in-the-wild videos~\citep{bahl2023vrb,chen2025vidbot}, and 2D point tracks or 3D object-point flow as foundation affordance~\citep{track2act,yuan2024general}.
A third group uses affordance to condition policies, including diffusion-policy and flow-matching action generators guided by 3D contact and post-contact trajectories~\citep{afforddp,anchordp3,afford_flow_matching}, hierarchical spatial-affordance plus low-level execution schemes~\citep{xu2025a0,rt_affordance}, and semantic 3D flow for generative control~\citep{g3flow}.
Rather than outputting discrete motion types, dense flow, hand/end-effector trajectories, or policy-conditioning signals, \ourmodel predicts a compact, object-centric 3D motion curve jointly with the functional mask.

\vspace{-3mm}
\paragraph{Affordance data pipelines and datasets.}
Affordance supervision can be categorized by annotation target, motion source, and labeling cost.
Directly annotated datasets cover early RGB-D part-affordance benchmarks~\citep{myers2015umd}, image-level functional grounding datasets~\citep{luo2022learning,ragnet,affordancer1,instructpart}, scene- and shape-level 3D functional annotations~\citep{deng20213daffordnet,delitzas2024scenefun3d}, robot-manipulation benchmarks~\citep{roboafford,roboafford_pp}, and human-video-derived datasets~\citep{gloverpp,heidinger2025_2handed,yoshida2024text}.
Beyond static affordance labels, trajectory and motion supervision is obtained from internet-video hand and wrist trajectories~\citep{bahl2023vrb,chen2025vidbot}, egocentric 6-DoF object trajectories with action descriptions~\citep{yoshida2024text,yoshida2025_6dof,egoscaler}, hand-object pose tracking datasets~\citep{hot3d}, scene-level 3D motion benchmarks~\citep{delitzas2024scenefun3d}, and simulated articulated-object interactions~\citep{mo2021where2act,wu2022vatmart,flowbot3d,flowbotpp}.
To reduce labeling cost, automatic or weakly supervised pipelines derive affordance labels from foundation-model distillation without dense annotation~\citep{tang2025uad}, egocentric-video affordance extraction~\citep{li2024precise,heidinger2025_2handed}, large-scale human-behavior mining~\citep{gloverpp}, part-prior weak supervision~\citep{xu2025partprior}, generative-AI augmentation for VLM affordance learning~\citep{roboafford_pp}, and MLLM-assisted grounding from human-object-interaction videos~\citep{videoafford,vagnet}.
Despite these efforts, existing resources remain limited in scale, especially for 3D motion supervision; \ourmodel{} addresses this gap with an extensible pipeline that aggregates one of the largest motion-affordance datasets to date.

\vspace{-2mm}
\section{Data Pipeline for \ourmodel}
\vspace{-1mm}
\label{sec:data}

\begin{figure}[t]
  \centering
    \includegraphics[width=\linewidth]{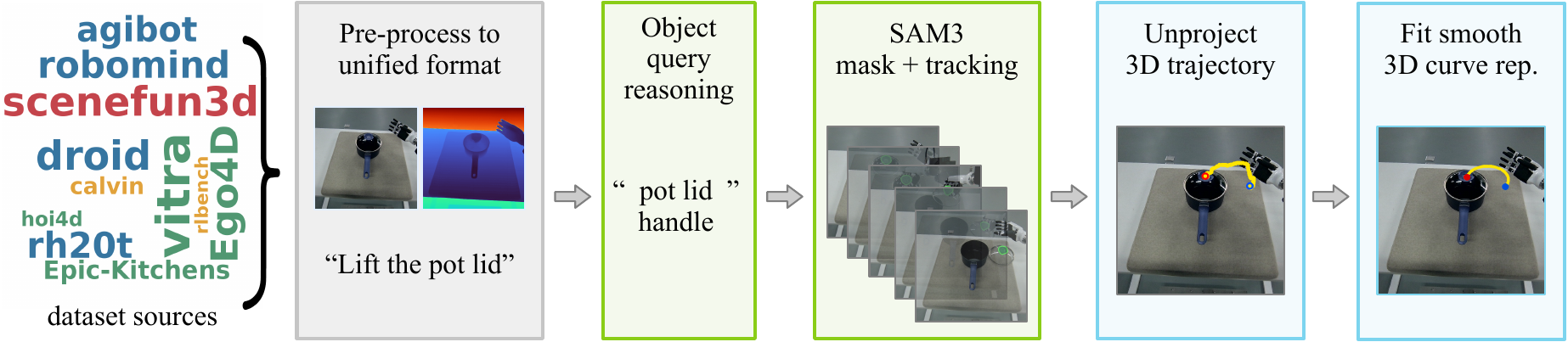}
  \caption{
    \textbf{Unified data collection pipeline.}
    We first aggregate data from various sources into a unified format (gray), then use Qwen3-VL~\citep{Qwen3-VL} and SAM3~\citep{sam3} to generate a functional affordance mask and a 2D tracking (green), and finally back-project them to obtain a 3D trajectory, which can be fit to a \curverep (blue). This scalable pipeline yields standardized, high-fidelity annotations with RGB-D observation, text phrase, mask, and 3D motion curve for training. (Best viewed in color.)
  }
  \label{fig:data_pipeline}
\end{figure}

Open-world affordance learning requires a large-scale dataset that covers diverse scenarios, tasks, objects, and action sequences while providing ground truth on \emph{what} to manipulate (segmentation) and \emph{how} to manipulate (motion). Existing datasets are either too small or only contain part of the information. In this paper, we build a unified data pipeline (Figure~\ref{fig:data_pipeline}) and curate a wide range of publicly available data, including robot demonstrations, egocentric human videos, and simulated interactions. We annotate all the data with a common affordance schema. Each data sample contains an RGB-D observation with a task description, a functional affordance mask, and a compact 3D motion trajectory.

\vspace{-3mm}
\paragraph{Dataset Curation.}

We curate datasets whose videos capture object interactions for functional purposes and have visible action regions and object motions.
Based on these criteria, we gathered 321{,}190 raw videos from 10 public sources, spanning human demonstrations~\citep{vitra,ego4d,rh20t,epickitech}, robot demonstrations~\citep{droid,rh20t,agiworld,robomind, robomind2}, simulation data~\citep{calvin,rlbench}, and real-world scans~\citep{delitzas2024scenefun3d}. Since a recording may contain multiple actions, we split the video episodes into action intervals, resulting in 1{,}242{,}740 intervals to start with.
Such broad source data pool provides us with diverse object categories, camera viewpoints, interaction tasks, and embodiments to construct our dataset. Further details are provided in Appendix~\ref{sec:appendix-data-statistics}.

\vspace{-3mm}
\paragraph{Dataset Preprocessing.}
To annotate at scale, we first preprocess all the above datasets into a common interval-based format, as illustrated in Fig.~\ref{fig:data_pipeline}-Gray.
 For each video episode in the datasets, we first decompose action intervals with existing annotations or per-dataset heuristics. Then for each action, we extract standardized schema, consisting of an input observation RGB-D frame, task language, camera parameters, and corresponding interval video clips.
Dataset-specific adapters also resolves raw storage formats and normalizes cameras. For moving-camera videos, we use camera poses to express tracked points in the camera canonical coordinate space of the observation frame.
We also use monocular depth estimators~\citep{lingbot,lin2025depthanything3} to improve depth quality. The dataset pre-processing steps are generally dataset-dependent. Further details are in Appendix~\ref{sec:appendix-data-preprocessing}.

\begin{wrapfigure}[13]{r}{0.48\linewidth}
  \centering
  \vspace{-3.5mm}
  \includegraphics[width=\linewidth]{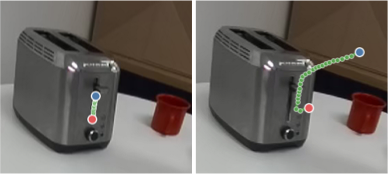}
  \vspace{-6mm}
  \caption{\footnotesize \textbf{Object trajectory vs.\ gripper heuristics.} Prior datasets often use hand or gripper trajectories as motion heuristics, but these can involve unwanted pre-contact motion (right). We track the object motion itself, which is more straightforward (left).}
  \label{fig:traj_compare}
\end{wrapfigure}

\vspace{-3mm}
\paragraph{Annotating Object Tracks and Masks.}
Prior works often use hand or gripper trajectories as the motion signal for affordance. However, this can entangle the undesired pre-contact hand motion with affordance-relevant post-contact object motion, as shown in Figure~\ref{fig:traj_compare} (right).
Instead, we use the tracking of the object as the post-contact motion in our affordance foundation model (Figure~\ref{fig:traj_compare}, left), as it directly indicates how the object moves after contact from a robot or a human.
To obtain the tracking, we first use a vision-language model to generate a short manipulable-part query from the task instruction and the observation/contact frames, then use SAM3~\citep{sam3} to track the manipulated object across the action interval (Figure~\ref{fig:data_pipeline}-Green). This step produces an object-centric motion trajectory and a functional affordance mask.

\vspace{-3mm}
\paragraph{Optimizing 3D Motion Curves.}
With the object mask and tracking trajectory, we recover the object's 3D motion trajectory by back-projecting the tracked masks and taking the mean of the 3D points positions in each frame. The resulting discrete path, however, is typically non-uniformly sampled and exhibits noise due to depth estimation errors and tracking inconsistencies. To address this, we fit a smooth parametric curve and convert it into our final canonical motion representation (\curverep) for training. The process is outlined in Fig.~\ref{fig:data_pipeline}-blue, with more detail in Appendix~\ref{sec:appendix-curvefit}.

\vspace{-3mm}
\paragraph{Filtering and Dataset Statistics.}
Before filtering, our data pool contains 1,242,740 action intervals
spanning robot teleoperation, human egocentric recordings, simulation, and real-world scans. At each step of processing, we filter low-quality clips, such as those with poor task grounding, occlusion, unreliable segmentation, and insufficient motion.
Each step removes around 1/2 of the samples, eventually resulting in 223,334 samples with valid motion labels.
We then perform manual annotation and quality control, and retain 59{,}867 training samples for \ourmodel. A dataset at this scale exposes the model to diverse interaction types, object categories, camera viewpoints, and embodiments.

\vspace{-2mm}
\section{Method}
\vspace{-1mm}
\label{sec:method}

\begin{figure*}[t]
\includegraphics[width=\linewidth]{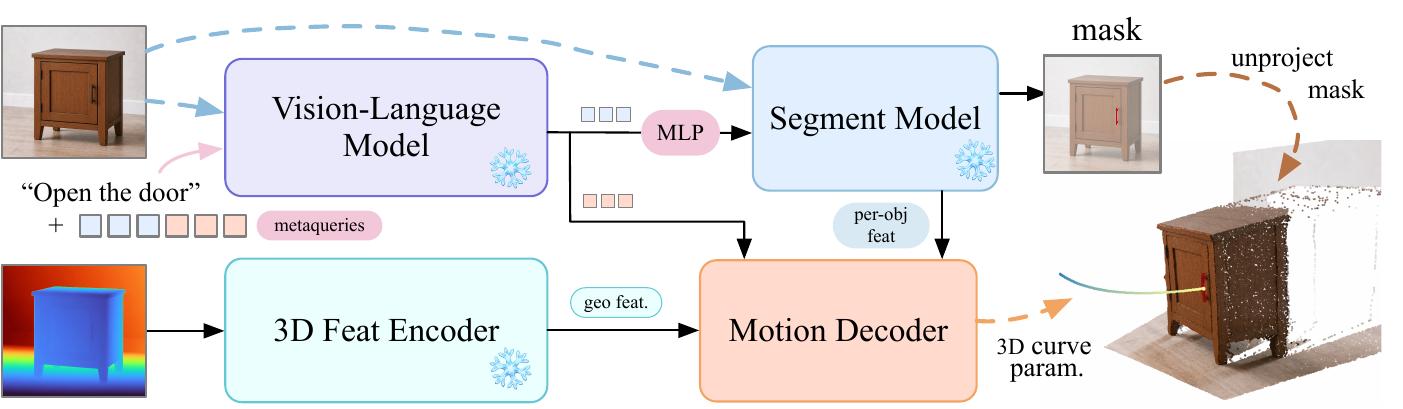}
  \caption{
    \textbf{\ourmodel architecture.}
    Starting from an RGB-D input and a task prompt,
    a frozen Qwen VLM encodes the language instruction into \textcolor{semanticblue}{\emph{semantic} tokens} and \textcolor{motionorange}{\emph{motion} tokens}, and a 3D encoder converts the depth observation into geometric features. With the language information encoded, the segmentation model generates the affordance segmentation mask from the RGB, while the motion decoder takes 3D features, task-conditioned context, and per-object features to produce a relative 3D motion prediction. Together, the mask and trajectory form the final deployable 3D affordance prediction. (Best view in color.)
  }
  \label{fig:pipeline}
\end{figure*}

\ourmodel takes an RGB-D observation and a task phrase as input, and jointly predicts a task-conditioned functional segmentation mask, along with a 3D post-contact motion curve in a single forward pass.
As shown in Fig.~\ref{fig:pipeline}, our model uses a simple architecture with two main components. First, we leverage the MetaQuery mechanism~\citep{pan2025metaquery} to connect a frozen Vision-Language Model (Qwen3-VL~\citep{Qwen3-VL}) with a segmentation model (SAM3~\cite{sam3}) to predict functional masks. Second, we use a 3D feature encoder~\citep{sonata} and a transformer decoder to predict 3D post-contact motion, represented as a \curverep. We describe the detailed network architecture in \S\ref{sec:method_arch} and the training scheme in \S\ref{sec:method_training}.

\vspace{-1mm}
\subsection{Network Architecture}
\label{sec:method_arch}

\paragraph{MetaQuery Conditioning.}
\label{sec:method_metaquery}
Introduced by \citet{pan2025metaquery}, MetaQuery serves as an interface to connect a frozen VLM with downstream models. In particular, a small set of learnable special tokens is appended to the VLM's input prompt and processed through the transformer. The hidden states in the final layer of the VLM serve as a compact conditioning feature for the downstream model.
\citet{pan2025metaquery} show that this approach can extract detailed visual conditions and transfer reasoning capabilities to multimodal generation tasks, such as image editing.

We bring the MetaQuery approach to our affordance prediction model, incorporating the reasoning capabilities from the VLM for functional mask segmentation and motion understanding. Specifically, we maintain two sets of learnable tokens:
$\mqsemset=\{\mqsem{0}, \dots, \mqsem{N_s-1}\}$, $\mqmotset=\{\mqmot{0}, \dots, \mqmot{N_m-1}\}$, where the first set $\mqsemset$ connects the VLM with the segmentation model, and the second $\mqmotset$ connects it with the motion prediction model.
The two sets of learnable tokens are appended to the input prompt together and processed through the transformer in Qwen3-VL~\cite{Qwen3-VL}. The last hidden states of each set of tokens are then fed into the downstream segmentation and motion model, respectively.
This joint formulation allows both the segmentation and motion models to share reasoning capabilities from VLMs within a single forward pass.

\vspace{-3mm}
\paragraph{Segmentation and Motion Decoding.}
\label{sec:method_seg}
With the MetaQuery tokens from VLMs, we predict the functional segmentation mask and 3D post-contact motion. For segmentation prediction, we primarily use SAM3~\citep{sam3}. Specifically, the semantic MetaQuery tokens $\mqsemset$ are first mapped by a two-layer MLP into SAM3's language-feature space. They are then passed through SAM3's mask decoder, which predicts per-detection boxes, masks, and object query features that are used for motion prediction.
By leveraging pretrained Qwen3-VL and SAM3, our model inherits the prior knowledge learned from large-scale pretraining for functional segmentation understanding.
For motion prediction, we additionally encode the point cloud (unprojected from the depth input) with a pretrained Sonata~\cite{sonata} network to provide 3D information, then project it back to images space and pool to geo features.
Afterwards, the motion decoder, which is a transformer decoder with self-attention to the encoded geo features and cross-attention to the per-object features from SAM3 and the motion MetaQuery tokens $\mqmotset$, to predict the parameters of motion curves below.

\vspace{-3mm}
\paragraph{Curved Motion Representation.}
\label{sec:method_motion}
A motion representation for open-world affordance must be expressive enough for complex interactions yet structured enough for robust manipulation. We therefore represent post-contact motion as an anchored 3D \curverep, parameterized by control points. The centroid of the masked depth map defines the start point \(\vP_0\), and the motion decoder predicts the remaining \(K\) ordered control points \(\{\vP_k\}_{k=1}^{K}\) in relative 3D coordinates. The trajectory is then computed with the Bernstein polynomial basis:
\begin{equation}
\vB(t) \;=\; \sum_{k=0}^{K} \binom{K}{k} (1-t)^{K-k}\, t^{k}\, \vP_k, \qquad t \in [0,1],
\label{eq:bezier}
\end{equation}

where \(\vB(t)\) is the 3D position at normalized time \(t\). The starting point \(\vP_0\) anchors the curve at the contact centroid, while the predicted control points parameterize the overall shape of the curve. Uniformly sampling \(t\in[0,1]\) produces executable 3D waypoints for robots.

\vspace{-1mm}
\subsection{Training Scheme}

\label{sec:method_training}

Directly training the full model is unstable: randomly initialized MetaQuery tokens provide a poor conditioning signal for SAM3, and noisy mask predictions would in turn make motion supervision ambiguous.
We therefore train our model in three stages: (I) aligning the MetaQuery interface with SAM3, (II) learning reliable task-conditioned affordance segmentation, and (III) fine-tuning motion prediction when the model is already robust in segmentation prediction. The pretrained priors, Qwen-VL, SAM3, and Sonata, are kept frozen throughout the training.

\paragraph{Stage 1: MetaQuery--SAM3 Alignment.}
Prior to end-to-end training, we initialize and train the MetaQuery tokens and projection MLP by aligning Qwen-derived features with SAM3's native text-conditioning space on the Visual Genome dataset~\citep{krishna2017visualgenome}.
For each caption-image pair, we encode the caption with SAM3's text encoder; in parallel, Qwen3-VL processes the same caption and image, and the projection MLP projects the resulting MetaQuery features into SAM3 text space.
We then run the SAM3 decoder with cross-attention to both the projected MetaQuery features and the original SAM3 text features. The decoder hidden states from the two branches are optimized with a Mean-Squared Error (MSE) loss, which provides a more stable initialization than training the new tokens directly from mask supervision.
This alignment step yields a strong initialization for the MetaQuery tokens and the Qwen-to-SAM3 MLP, thereby stabilizing subsequent joint affordance training.

\vspace{-3mm}
\paragraph{Stage 2: End-to-End Training for Affordance Segmentation.}
In the second stage, we train our affordance segmentation model end-to-end on an aggregated mixture of four affordance datasets: HOVA-500K~\citep{gloverpp}, RAGNet~\citep{ragnet}, InstructPart~\citep{instructpart}, and ReasonAFF~\citep{affordancer1}.
The unfrozen parameters are identical to those in Stage 1: the MetaQuery tokens and the projection MLP.
The motion prediction branch is disabled, and we only train the model with objectives from SAM3~\cite{sam3}, which combines Hungarian-matched box regression ($\ell_1$ + GIoU), presence classification, per-query mask prediction (focal BCE + Dice), and a semantic-segmentation term (focal + Dice + presence), all averaged with the same hyperparameters as SAM3. We refer readers to the original paper for more details.

\vspace{-3mm}
\paragraph{Stage 3: Joint Motion and Segmentation Training.}
In the final stage, we train segmentation and motion prediction jointly on our own aggregated affordance dataset curated from Section~\ref{sec:data}, together with the Stage~2 training data. The total objective combines the Stage~2 SAM3 grounding loss $\lossSam$, down-weighted to prevent the segmentation head from overfitting, with a curve loss $\lossCurve$ on sampled trajectory points to learn the motion:
\begin{equation}
\loss \;=\; \wSam\, \lossSam \;+\; \wCurve\, \lossCurve.
\label{eq:total_loss}
\end{equation}
We follow the point-sampling loss from Curve-GCN~\citep{ling2019curvegcn} to supervise motion prediction. Specifically, for each SAM3-matched query $(b,q) \in \matchset$ returned by the Hungarian matcher, we evaluate both the predicted B\'ezier curve $\pred{\vB}_{b,q}(t)$ and its matched ground-truth curve $\gt{\vB}_{b,q}(t)$ on a fixed uniform time interval $\{t_i = i/(T-1)\}_{i=0}^{T-1}$ and minimize the $\ell_1$ distance between the sampled points,
\begin{equation}
\lossCurve \;=\; \frac{1}{|\matchset|\,T} \sum_{(b,q)\in\matchset} \sum_{i=0}^{T-1} \bigl\| \pred{\vB}_{b,q}(t_i) - \gt{\vB}_{b,q}(t_i) \bigr\|_{1}.
\label{eq:curve_loss}
\end{equation}
In practice, we find this point-sampling supervision substantially more effective than directly regressing the locations of control points.

\vspace{-2mm}
\section{Experiments}
\label{sec:experiments}

\vspace{-2mm}
\subsection{Implementation Details}
\label{sec:exp_implementation}

We use Qwen3-VL-8B~\citep{Qwen3-VL} as our VLM backbone, SAM3~\citep{sam3} as the segmentation model, and Sonata~\citep{sonata} as the 3D feature encoder; all three pretrained components are frozen throughout training. The motion decoder uses six transformer layers. Along with the MLPs and MetaQuery tokens, our model adds only \textbf{32.21M} trainable parameters on top of the pretrained models. We use 64 MetaQuery tokens in total, where each semantic and motion branch has 32. We set $\lambda_{\mathrm{SAM}}=0.5$, $\lambda_{\mathrm{curve}}=100$, and train the model with a learning rate of $2\times 10^{-4}$. For the point-sampling loss, we sample $T=16$ points per curve. We train \ourmodel{} on \textbf{$4\times$ NVIDIA GH200} GPUs for approximately eight days. The three stages mentioned above use batch sizes of 196, 128, and 96, respectively, and run for 10{,}000, 40{,}000, and 20{,}000 steps, respectively.

\vspace{-1mm}
\subsection{Affordance Evaluation}
\label{sec:exp_seg}

To comprehensively demonstrate the affordance understanding capability of \ourmodel, we evaluate it from three perspectives: accuracy of affordance mask segmentation, correctness of the contact point derived from the mask, and quality of 3D motion. 

\vspace{-2mm}
\subsubsection{Affordance Segmentation Evaluation}
\vspace{-1mm}
We first evaluate \ourmodel's capability to reason about \emph{where} the affordance lies by measuring segmentation quality. 
We compare against three baselines: a zero-shot Qwen3-VL-8B~\citep{Qwen3-VL} object query generation + SAM3~\citep{sam3} mask generation pipeline, AffordanceNet~\citep{ragnet}, and Affordance-R1~\citep{affordancer1}. 

\begin{figure}[t]
  \centering
  \includegraphics[width=\linewidth]{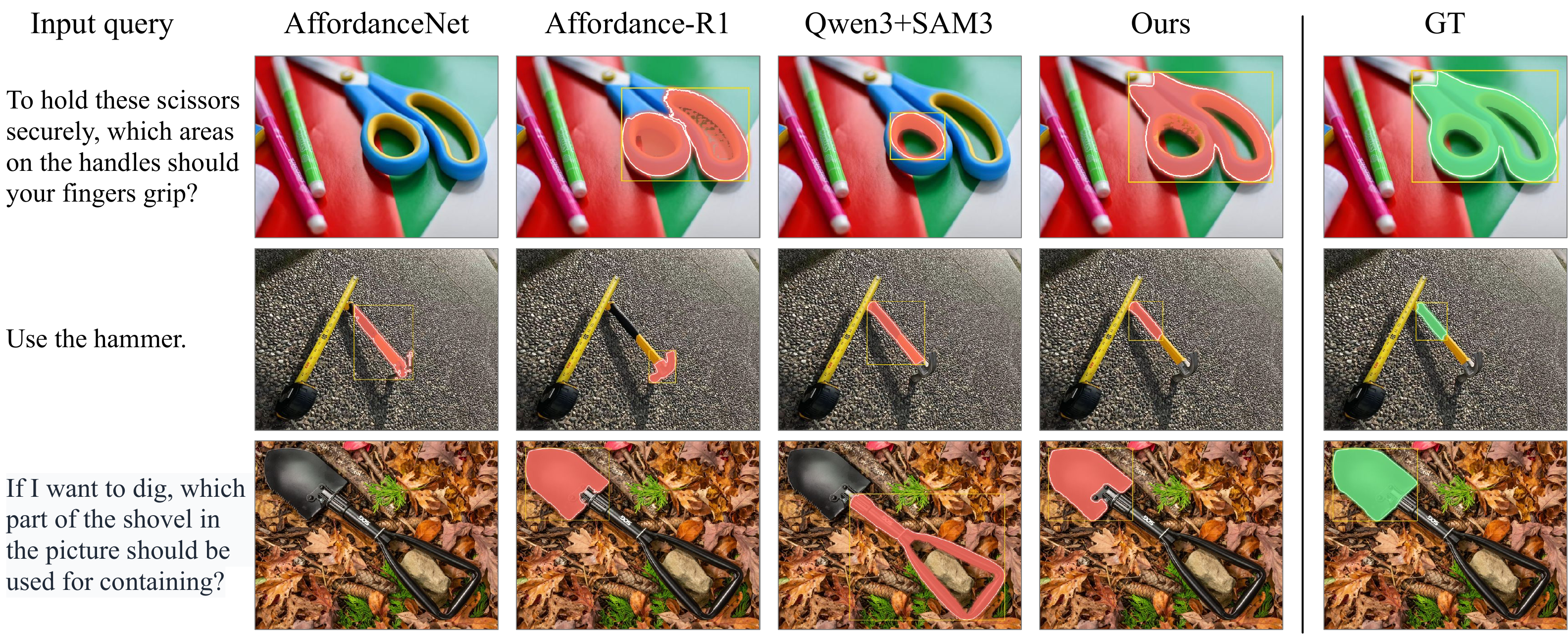}
  \caption{\textbf{Qualitative Examples on Affordance Segmentation.}
  \ourmodel{} accurately segments task-specific affordance regions, including complex scissor handles with holes and intent-dependent regions such as shovel blades for containing or hammer handles for using. We provide more examples in Appendix~\ref{sec:appendix-ourmodeltestset-gallery}.}
  \label{fig:quali_seg}
\end{figure}

We show qualitative task-conditioned affordance mask segmentation results in Fig.~\ref{fig:quali_seg}. \ourmodel{} consistently predicts the correct affordance region for the diverse task instructions, can precisely segment complex regions (scissors handle with holes), and strictly aligns with the given task (shovel blade--containing, hammer handle--using), demonstrating superior performance in reasoning about the task-specific affordance compared to baselines.

Quantitatively, following our baselines~\citep{ragnet,gloverpp,affordancer1}, we evaluate on eight test sets drawn from four affordance benchmarks, and report gIoU and cIoU metrics in Table~\ref{tab:seg_results}.
Across all test sets, \ourmodel{} outperforms all the baselines, achieves the best gIoU and cIoU, and improves overall mean gIoU/cIoU by $23.9/26.3$ points over the strongest baseline.
Notably, even when using the Qwen3-VL-2B variant with fewer parameters, \ourmodel{} remains superior to the baseline models by a large margin.

\begin{table*}[!h]
\centering
\vspace{-5mm}
\caption{Quantitative results on affordance mask segmentation (gIoU / cIoU, \%; higher is better). Best results are \textbf{bolded}; \ourmodel significantly outperforms all the baselines across 8 datasets in both metrics. Even with a smaller 2B model, \ourmodel still improves over the baselines by a large margin.}
\label{tab:seg_results}
\resizebox{\textwidth}{!}{%
\begin{tabular}{l c c c c c c c c c}
\toprule
Method
& HANDAL-Mini & 3DOI & HANDAL Easy & HANDAL Hard & 3DOI Easy
& HOVA-500K & ReasonAFF & InstructPart & Mean \\
\midrule
Qwen3-VL-8B~\citep{Qwen3-VL} + SAM3~\citep{sam3} & 40.3 / 44.4 & 36.4 / 22.5 & 44.8 / 50.6 & 45.2 / 52.2 & 44.3 / 27.2 & 63.3 / 38.5 & 31.9 / 19.6 & 41.4 / 28.1 & 42.5 / 36.5 \\
Affordance-R1~\citep{affordancer1} & 18.2 / 10.6 & 51.8 / 38.5 & 39.8 / 32.0 & 37.4 / 26.5 & 61.4 / 53.6 & 27.9 / 14.1 & 67.1 / 62.1 & 67.2 / 58.6 & 45.4 / 36.2 \\
AffordanceNet~\citep{ragnet}       & 59.6 / 58.8 & 39.9 / 37.8 & 55.2 / 50.7 & 52.8 / 50.3 & 36.3 / 36.6 & 51.7 / 28.1 & 25.4 / 19.9 & 29.0 / 21.8 & 45.0 / 40.9 \\
\midrule
\ourmodel (Ours)                  & \textbf{60.3} / 59.9 & \textbf{58.9} / \textbf{56.0} & \textbf{67.7} / \textbf{68.8} & \textbf{68.9} / \textbf{69.2} & 75.5 / 72.6 & \textbf{80.8} / 53.9 & \textbf{78.1} / \textbf{78.2} & \textbf{78.3} / \textbf{81.4} & \textbf{69.3} / \textbf{67.2} \\
\ourmodel-2B                      & \textbf{60.3} / \textbf{60.3} & 56.0 / 37.9 & 66.0 / 66.8 & 66.5 / 67.1 & \textbf{75.8} / \textbf{76.8} & \textbf{80.8} / \textbf{56.1} & 71.9 / 73.5 & 65.8 / 54.1 & 66.6 / 61.8 \\
\midrule
\end{tabular}%
}
\vspace{-3mm}
\end{table*}

\begin{table*}[h]
\centering
\vspace{-2mm}
\caption{Contact point evaluation with point hit rate (\%; higher is better). We compare with 2D point-based methods. Best per column in \textbf{bold}. We use the Pole of Inaccessibility of the predicted mask as the predicted point.}
\label{tab:point_hitrate}
\resizebox{\textwidth}{!}{%
\begin{tabular}{l cccccccc}
\toprule
Method & HANDAL-Mini & 3DOI & HANDAL Easy & HANDAL Hard & 3DOI Easy & HOVA-500K & ReasonAFF & InstructPart \\
\midrule
A0~\citep{xu2025a0}    &           4.6 &          3.6 &           5.0 &           6.0 &          3.6 &           3.8 &          20.7 &          22.7 \\
VRB~\citep{bahl2023vrb}   &          31.5 &         46.6 &          31.5 &          31.3 &         48.4 &          22.4 &          35.2 &          39.8 \\
GLOVER++~\citep{gloverpp} &          67.6 &          6.8 &          39.2 &          34.4 &          4.9 &          28.6 &           4.3 &           4.7 \\
\midrule
\ourmodel (Ours)         & \textbf{80.3} & \textbf{67.3} & \textbf{88.2} & \textbf{88.8} & \textbf{82.6} & \textbf{88.3} & \textbf{96.5} & \textbf{95.5} \\
\bottomrule
\end{tabular}%
}

\end{table*}

\vspace{-3mm}
\subsubsection{Contact Point Evaluation} 
\vspace{-1mm}
Beyond using masks for affordance, prior work also adopts contact points as an affordance representation; we therefore compare with A0~\citep{xu2025a0}, GLOVER++~\citep{gloverpp}, VRB~\citep{bahl2023vrb}, and measure whether a predicted contact point lies on the ground-truth affordance mask. 
For \ourmodel, we take the Pole of Inaccessibility~\citep{GarciaCastellanos2007} of the predicted mask as the contact point.
We use hit rate $\Pr[\text{point}\!\in\!\text{GT mask}]$, which measures whether the predicted contact point lies on the affordance mask as the evaluation metric. As shown in Table~\ref{tab:point_hitrate}, \ourmodel significantly outperforms the best baseline by $12.7\%$--$61.3\%$ ($55.7\%$ on InstructPart and $61.3\%$ on ReasonAFF).

\begin{figure}[!t]
  \centering
  \includegraphics[width=\linewidth]{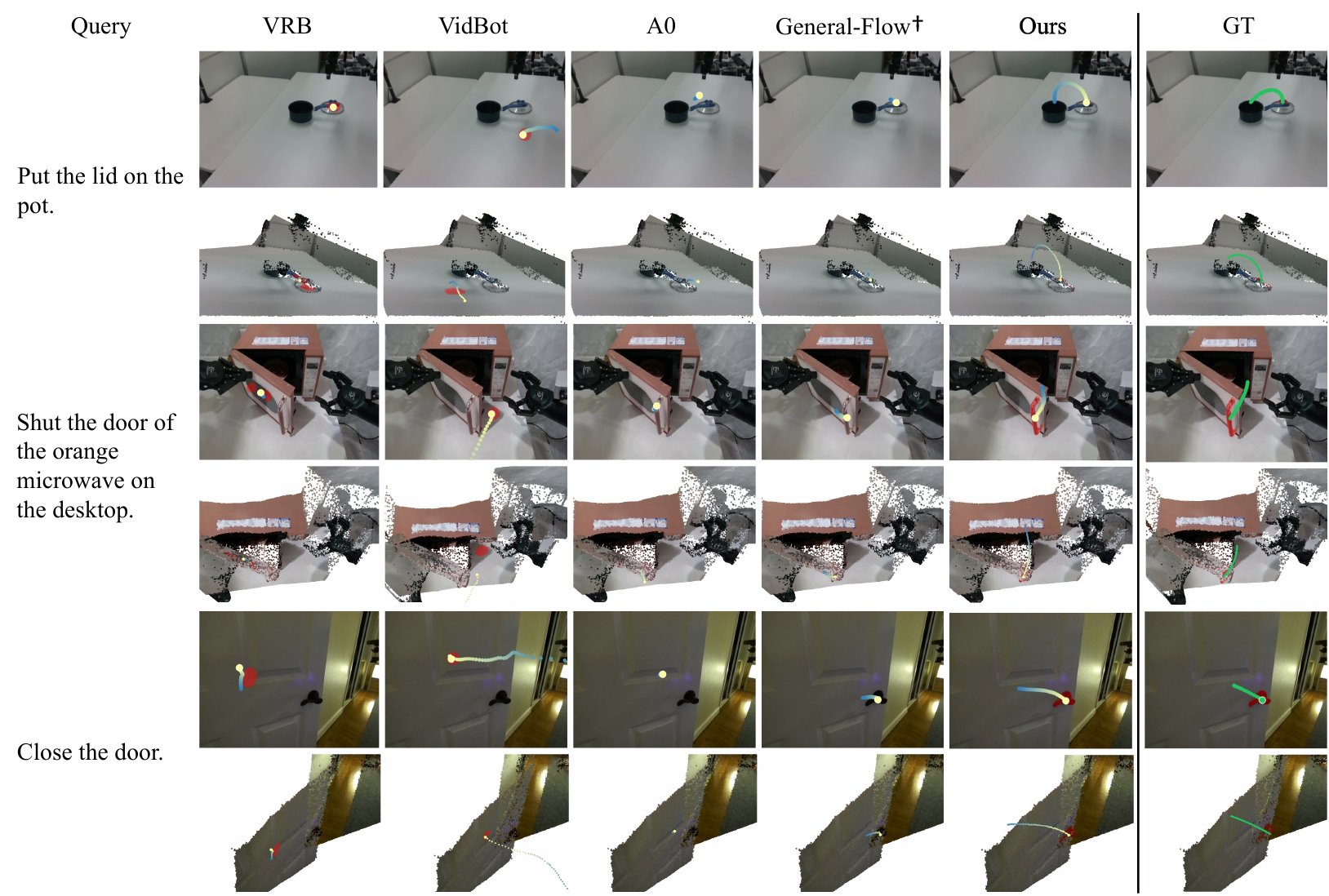}
  \caption{\textbf{Qualitative motion prediction results.}
  \ourmodel{} accurately localizes the actionable object region and predicts smooth, task-aligned 3D motion curves, whereas the baselines often fail to identify the relevant affordance region or produce physically plausible motion. $^\dagger$ General~Flow used the mask prediction from our \ourmodel{} for its starting query points.}
  \label{fig:qualitative}
\end{figure}

\vspace{-2mm}
\subsubsection{3D Motion Evaluation} 

\label{sec:exp_motion3d}

\vspace{-1mm}
\paragraph{Evaluation Datasets.}
We evaluate $3$D motion on three test sets with different domain shifts. (I) The \ourmodel{} test set ($121$ examples) is a cross-source split randomly sampled from the high-quality set we curated. This test set is further verified through a second human quality-control pass, and we exclude these data samples from the training set to prevent data leak. 
(II)  The SceneFun3D~\citep{delitzas2024scenefun3d} test set ($721$ examples) comes from the original validation set in SceneFun3D and contains scenes that are not present in the training set. For each task in each scene, we use the first frame in which the target object is visible for evaluation (details of dataset processing in Appendix~\ref{sec:appendix-data-preprocessing}). 
(III) The RoboMIND2 dataset test set ($156$ examples) is an out-of-domain test set deliberately excluded from training. We keep functionality-related tasks and remove relocation-only instructions such as ``place A to B'' for evaluation, as such waypaths are usually non-deterministic.
\paragraph{Evaluation Metrics and Baselines.}
We evaluate predicted 3D motion curves using Average Displacement Error (ADE), Final Displacement Error (FDE)~\citep{liu2022joint,yuan2024general} computed in both absolute scale and relative scale,
contact-in-mask hit rate (CIM).
We compare \ourmodel{} with four 3D affordance baselines: A0~\citep{xu2025a0}, VRB~\citep{bahl2023vrb}, VidBot~\citep{chen2025vidbot}, and General~Flow~\citep{yuan2024general}.
For each baseline, we follow their official protocol to obtain the 3D motion predictions, and linearly interpolate every prediction and every ground-truth trajectory to a common length of $T{=}50$ points for evaluation. 
Note that General~Flow~\citep{yuan2024general}  requires manually-specified query points for motion prediction, so we ``lend'' the predicted mask from our model for its query sampling. 

\begin{table}[t]
\centering
\caption{Quantitative 3D motion evaluation. ADE/FDE are in meters; subscript $a$ is absolute, $r$ is relative. CIM is the contact-in-mask hit rate. Best per dataset in \textbf{bold}. General Flow$^\dagger$ gives no starting point $\mathbf{r}_0$ for motion prediction, and uses predicted mask from \ourmodel to get its query points. Yet, it still underperforms our model.}
\label{tab:motion_3d}
\resizebox{\columnwidth}{!}{%
\begin{tabular}{ll|cc cc cc}
\toprule
Dataset & Method
 & ADE$_a\!\downarrow$ & FDE$_a\!\downarrow$
 & ADE$_r\!\downarrow$ & FDE$_r\!\downarrow$
 & CIM\,\%$\uparrow$ & \#fail$\downarrow$ \\
\midrule
\multirow{5}{*}{\shortstack[l]{\ourmodeltestset{}\\($n{=}121$)}}
 & A0~\citep{xu2025a0}           & 0.378 & 0.369 & 0.140 & 0.284 &  6.6 & \textbf{0} \\
 & VRB~\citep{bahl2023vrb}          & 0.242 & 0.297 & 0.087 & 0.220 & 11.0 & 3 \\
 & VidBot~\citep{chen2025vidbot}    & 0.520 & 0.614 & 0.192 & 0.364 &  6.7 & 2 \\
 & General~Flow$^\dagger$~\citep{yuan2024general} & 0.110 & 0.230 & 0.088 & 0.225 & -- & \textbf{0} \\
 & \ourmodel (Ours)                & \textbf{0.098} & \textbf{0.139} & \textbf{0.080} & \textbf{0.135} & \textbf{81.0} & \textbf{0} \\
\midrule
\multirow{5}{*}{\shortstack[l]{SceneFun3D test\\($n{=}721$)}}
 & A0~\citep{xu2025a0}           & 1.008 & 1.066 & 0.249 & 0.476 &  3.3 & \textbf{0} \\
 & VRB~\citep{bahl2023vrb}          & 0.606 & 0.702 & 0.227 & 0.446 & 11.8 & 36 \\
 & VidBot~\citep{chen2025vidbot}    & 0.772 & 0.848 & 0.201 & 0.393 &  9.4 & 11 \\
 & General~Flow$^\dagger$~\citep{yuan2024general} & 0.413 & 0.572 & 0.212 & 0.413 & -- & 4 \\
 & \ourmodel (Ours)                & \textbf{0.351} & \textbf{0.441} & \textbf{0.135} & \textbf{0.260} & \textbf{67.3} & 1 \\
\midrule
\multirow{5}{*}{\shortstack[l]{RoboMIND2 test\\($n{=}156$)}}
 & A0~\citep{xu2025a0}           & 0.374 & 0.388 & 0.193 & 0.327 &  3.2 & \textbf{0} \\
 & VRB~\citep{bahl2023vrb}          & 0.299 & 0.373 & 0.190 & 0.330 & 27.4 & 10 \\
 & VidBot~\citep{chen2025vidbot}    & 0.368 & 0.479 & 0.240 & 0.414 & 23.7 & 5 \\
 & General~Flow$^\dagger$~\citep{yuan2024general} & 0.260 & 0.369 & 0.184 & 0.314 & -- & 2 \\
 & \ourmodel (Ours)                & \textbf{0.254} & \textbf{0.323} & \textbf{0.177} & \textbf{0.276} & \textbf{62.2} & \textbf{0} \\
\bottomrule
\end{tabular}%
}
\end{table}

\paragraph{Evaluation Results.}
We provide quantitative results in Table~\ref{tab:motion_3d} and qualitative results in Fig.~\ref{fig:qualitative}.
\ourmodel{} achieves the best ADE and FDE in both absolute and relative scale on all three test sets, and significantly outperforms the baselines in CIM. 
Even when General~Flow is evaluated under a favorable protocol, as it is provided with \ourmodel{}'s predicted mask and start anchor, \ourmodel{} still achieves substantially better motion prediction results. This advantage is further shown in Fig.~\ref{fig:qualitative}: \ourmodel{} produces task-aligned masks and motions, whereas the baselines often produce both implausible object localization and task-inconsistent trajectories.

\vspace{-1mm}
\subsection{Ablations}
\label{sec:exp_ablations}

\begin{table}[t]
\centering
\begin{minipage}[t]{0.45\linewidth}
\centering
\caption{LLM backbone ablation.}
\label{tab:ablation-llm}
\vspace{-2mm}
\resizebox{\linewidth}{!}{%
\begin{tabular}{l cc}
\toprule
Variant & Mean gIoU$\uparrow$ & Mean cIoU$\uparrow$ \\
\midrule
\ourmodel (Ours) & \textbf{69.3} & \textbf{67.2} \\
w/ Qwen3-VL-2B  & 66.6 & 61.8 \\
w/ Qwen3.5-9B   & 61.8 & 54.4 \\
\bottomrule
\end{tabular}%
}
\vspace{-5mm}
\end{minipage}
\hfill
\begin{minipage}[t]{0.54\linewidth}
\centering
\caption{3D motion ablations on RoboMIND2.}
\label{tab:ablation-motion}
\vspace{-2mm}
\resizebox{\linewidth}{!}{%
\begin{tabular}{l cc}
\toprule
Variant & ADE$_a\downarrow$ & FDE$_a\downarrow$ \\
\midrule
\ourmodel (Ours) & \textbf{0.254} & \textbf{0.323} \\
w/ DFormerv2~\citep{dformer} (3D feat. encoder)       & 0.273 & 0.348 \\
w/ OPD~\citep{opd} (curve parameterization)      & 0.282 & 0.374 \\
\bottomrule
\end{tabular}%
}
\vspace{-5mm}
\end{minipage}
\end{table}

We ablate three different design choices of our model: the LLM backbone, 
the 3D feature encoder, and the motion curve parameterization. Results are provided in Table~\ref{tab:ablation-llm} and Table~\ref{tab:ablation-motion}. 

For different LLM backbones, we train the model using the same recipe as our default model and report the evaluation performance on all the 8 test sets. Our default model with Qwen3-VL-8B achieves the best segmentation performance, outperforming both the smaller model Qwen3-VL-2B and the larger Qwen3.5-9B. 
We hypothesize: the reason why larger Qwen3.5-9B underperforms is that its general-purpose MoE design might be less suited to dense vision--language prediction.

For the 3D feature encoder, we replace Sonata with the \emph{DFormerv2}~\citep{dformer} architecture and train it using the same recipe. We evaluate the performance on the open-domain RoboMIND2 test set. Our default 3D feature encoder outperforms \emph{DFormerv2}~\citep{dformer}, benefiting from stronger 3D geometric cues in point cloud-derived features. 

For motion curve representation, we compare our curve parameterization with the representation used in \emph{OPD}~\citep{opd}. Our representation achieves better results, as the single parameterization for multiple motion types in OPD can introduce ambiguity.

\vspace{-1mm}
\subsection{Real-Robot Demonstration}

\label{sec:exp_realrobot}

Deploying \ourmodel on real robotic platforms is straightforward and requires no additional task-specific heuristics. 
Given a calibrated RGB-D input from one camera, \ourmodel predicts a contact mask and post-contact motion trajectory; the mask is back-projected to localize the target object, while AnyGrasp~\cite{fang2023anygrasp} estimates feasible grasp poses from the reconstructed scene point cloud.
The predicted trajectory, represented as a smooth spline curve, provides a local tangent direction for adapting the gripper orientation, enabling rotational manipulation such as opening a microwave. This orientation-aware execution is difficult to obtain from line-based trajectory predictions in prior approaches~\citep{bahl2023vrb,xu2025a0,yuan2024general}. 

\begin{figure}[h]
  \centering
  \includegraphics[width=\linewidth]{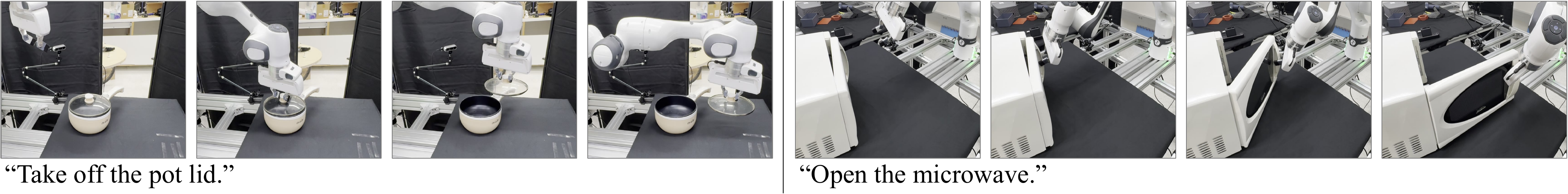}
  \caption{
    \textbf{Real-robot deployment (Franka).}
    \ourmodel{} can be directly deployed to a real robotic system without any additional task-specific heuristics. Given a task from the user, our model can accurately locate the actionable (grasping) region and produce an accurate post-contact trajectory for robot manipulation.
  }
  \label{fig:robotic_demo}
\end{figure}

\begin{wraptable}[6]{r}{0.33\linewidth}
  \vspace{-10mm}
  \centering
  \caption{Real-world Task Performance.}
  \vspace{+1mm}
  \label{tab:task_success_rates}
  \resizebox{\linewidth}{!}{%
  \begin{tabular}{l|c}
    \hline
    \textbf{Task} & \textbf{Success Rate} \\
    \hline
    Pick Up Screwdriver & 1.0 \\
    Take Off Pot Lid & 1.0 \\
    Open Drawer & 0.8 \\
    Open Microwave & 0.8 \\
    \hline
  \end{tabular}%
  }
\end{wraptable}
We evaluate \ourmodel on four real-world tasks: Pick Up Screwdriver, Take Off Pot Lid, Open Drawer, and Open Microwave, using a Franka Research 3 arm and two calibrated third-person RGB-D RealSense D435 cameras. For each task, \ourmodel uses one RGB-D observation as input, while observations from both cameras are fused into the scene point cloud used by AnyGrasp.
We report success rates in Tab.~\ref{tab:task_success_rates} and qualitative examples in Fig.~\ref{fig:robotic_demo}. \ourmodel achieves an average success rate of 90\%, demonstrating reliable real-robot deployment for both contact-centric grasping and orientation-aware articulated-object manipulation.

\vspace{-4mm}
\section{Conclusion}
\vspace{-1mm}
\label{sec:conclusion}

In this paper, we present \textbf{\ourmodel}, a step towards an affordance foundation model for understanding functionality.
From a single RGB-D observation and a language task description, \ourmodel predicts a task-conditional functional mask (\emph{where} to interact) and a 3D post-contact motion curve (\emph{how} to interact).
To achieve open-world generalization, we build a large-scale standardized data pipeline that converts heterogeneous robot, human, simulation, and real-world scan data into a shared affordance schema with language, masks, and object-centric 3D motion annotations.
Empirically, \ourmodel outperforms all baselines on affordance segmentation across eight test sets from four benchmarks; predicts substantially more accurate contact points; and achieves the best 3D motion performance on all three motion test sets.
Without embodiment-specific finetuning, \ourmodel can be directly deployed in the real robot for manipulation, suggesting a practical path towards open-world affordance models that unify functionality perception with executable action. We provide limitations, failure cases, and future directions in Appendix~\ref{sec:app_limit}.


\clearpage  
\bibliographystyle{plainnat}
\bibliography{references,sections/related_work/related_work}

@inproceedings{liu2022joint,
  title={Joint Hand Motion and Interaction Hotspots Prediction from Egocentric Videos},
  author={Liu, Shaowei and Tripathi, Subarna and Majumdar, Somdeb and Wang, Xiaolong},
  booktitle = {Proceedings of the IEEE/CVF Conference on Computer Vision and Pattern Recognition (CVPR)},
  year={2022}
}

@article{Qwen3-VL,
      title={Qwen3-VL Technical Report}, 
      author={Shuai Bai and Yuxuan Cai and Ruizhe Chen and Keqin Chen and Xionghui Chen and Zesen Cheng and Lianghao Deng and Wei Ding and Chang Gao and Chunjiang Ge and Wenbin Ge and Zhifang Guo and Qidong Huang and Jie Huang and Fei Huang and Binyuan Hui and Shutong Jiang and Zhaohai Li and Mingsheng Li and Mei Li and Kaixin Li and Zicheng Lin and Junyang Lin and Xuejing Liu and Jiawei Liu and Chenglong Liu and Yang Liu and Dayiheng Liu and Shixuan Liu and Dunjie Lu and Ruilin Luo and Chenxu Lv and Rui Men and Lingchen Meng and Xuancheng Ren and Xingzhang Ren and Sibo Song and Yuchong Sun and Jun Tang and Jianhong Tu and Jianqiang Wan and Peng Wang and Pengfei Wang and Qiuyue Wang and Yuxuan Wang and Tianbao Xie and Yiheng Xu and Haiyang Xu and Jin Xu and Zhibo Yang and Mingkun Yang and Jianxin Yang and An Yang and Bowen Yu and Fei Zhang and Hang Zhang and Xi Zhang and Bo Zheng and Humen Zhong and Jingren Zhou and Fan Zhou and Jing Zhou and Yuanzhi Zhu and Ke Zhu},
	  journal={arXiv preprint arXiv:2511.21631},
      year={2025}
}

@article{pan2025metaquery,
  title   = {Transfer between Modalities with {MetaQueries}},
  author  = {Pan, Xichen and Shukla, Satya Narayan and Singh, Aashu and Zhao, Zhuokai and Mishra, Shlok Kumar and Wang, Jialiang and Xu, Zhiyang and Chen, Jiuhai and Li, Kunpeng and Juefei-Xu, Felix and Hou, Ji and Xie, Saining},
  journal = {arXiv preprint arXiv:2504.06256},
  year    = {2025}
}

@book{gibson1979ecological,
  title={The Ecological Approach to Visual Perception},
  author={Gibson, James J.},
  year={1979},
  publisher={Houghton Mifflin}
}

@inproceedings{tang2025uad,
  title={{UAD}: Unsupervised Affordance Distillation for Generalization in Robotic Manipulation},
  author={Tang, Yihe and Huang, Wenlong and Wang, Yingke and Li, Chengshu and Yuan, Roy and Zhang, Ruohan and Wu, Jiajun and Fei-Fei, Li},
  booktitle={IEEE International Conference on Robotics and Automation (ICRA)},
  year={2025}
}

@inproceedings{yuan2024robopoint,
  author    = {Yuan, Wentao and Duan, Jiafei and Blukis, Valts and Pumacay, Wilbert and Krishna, Ranjay and Murali, Adithyavairavan and Mousavian, Arsalan and Fox, Dieter},
  title     = {{RoboPoint}: A Vision-Language Model for Spatial Affordance Prediction in Robotics},
  booktitle = {Proceedings of The 8th Conference on Robot Learning (CoRL)},
  series    = {Proceedings of Machine Learning Research},
  volume    = {270},
  pages     = {4005--4020},
  year      = {2024},
  publisher = {PMLR},
  url       = {https://proceedings.mlr.press/v270/yuan25c.html}
}

@inproceedings{xu2025a0,
  title={A0: An Affordance-Aware Hierarchical Model for General Robotic Manipulation},
  author={Xu, Rongtao and Zhang, Jian and Guo, Minghao and Wen, Youpeng and Yang, Haoting and Lin, Min and Huang, Jianzheng and Li, Zhe and Zhang, Kaidong and Wang, Liqiong and Kuang, Yuxuan and Cao, Meng and Zheng, Feng and Liang, Xiaodan},
  booktitle={Proceedings of the IEEE/CVF International Conference on Computer Vision (ICCV)},
  year={2025}
}

@inproceedings{bahl2023vrb,
  title={Affordances from Human Videos as a Versatile Representation for Robotics},
  author={Bahl, Shikhar and Mendonca, Russell and Chen, Lili and Jain, Unnat and Pathak, Deepak},
  booktitle={Proceedings of the IEEE/CVF Conference on Computer Vision and Pattern Recognition (CVPR)},
  year={2023}
}

@article{fang2023anygrasp,
  title={AnyGrasp: Robust and Efficient Grasp Perception in Spatial and Temporal Domains},
  author={Fang, Hao-Shu and Wang, Chenxi and Fang, Hongjie and Gou, Minghao and Liu, Jirong and Yan, Hengxu and Liu, Wenhai and Xie, Yichen and Lu, Cewu},
  journal={IEEE Transactions on Robotics},
  year={2023},
  doi={10.1109/TRO.2023.3281153}
}

@inproceedings{chen2025vidbot,
  author    = {Chen, Hanzhi and Sun, Boyang and Zhang, Anran and Pollefeys, Marc and Leutenegger, Stefan},
  title     = {{VidBot}: Learning Generalizable {3D} Actions from In-the-Wild {2D} Human Videos for Zero-Shot Robotic Manipulation},
  booktitle = {Proceedings of the IEEE/CVF Conference on Computer Vision and Pattern Recognition (CVPR)},
  pages     = {27661--27672},
  year      = {2025}
}

@inproceedings{ragnet,
  title     = {{RAGNet}: Large-scale Reasoning-based Affordance Segmentation Benchmark towards General Grasping},
  author    = {Wu, Dongming and Fu, Yanping and Huang, Saike and Liu, Yingfei and Jia, Fan and Liu, Nian and Dai, Feng and Wang, Tiancai and Anwer, Rao Muhammad and Khan, Fahad Shahbaz and Shen, Jianbing},
  booktitle = {Proceedings of the IEEE/CVF International Conference on Computer Vision (ICCV)},
  year      = {2025}
}

@inproceedings{instructpart,
  title     = {{InstructPart}: Task-Oriented Part Segmentation with Instruction Reasoning},
  author    = {Wan, Zifu and Xie, Yaqi and Zhang, Ce and Lin, Zhiqiu and Wang, Zihan and Stepputtis, Simon and Ramanan, Deva and Sycara, Katia},
  booktitle = {Proceedings of the 63rd Annual Meeting of the Association for Computational Linguistics (ACL)},
  year      = {2025}
}

@inproceedings{affordancer1,
  title     = {{Affordance-R1}: Reinforcement Learning for Generalizable Affordance Reasoning in Multimodal Large Language Models},
  author    = {Wang, Hanqing and Wang, Shaoyang and Zhong, Yiming and Yang, Zemin and Wang, Jiamin and Cui, Zhiqing and Yuan, Jiahao and Han, Yifan and Liu, Mingyu and Ma, Yuexin},
  booktitle = {Proceedings of the AAAI Conference on Artificial Intelligence (AAAI)},
  year      = {2026}
}

@inproceedings{yuan2024general,
  author    = {Yuan, Chengbo and Wen, Chuan and Zhang, Tong and Gao, Yang},
  title     = {General Flow as Foundation Affordance for Scalable Robot Learning},
  booktitle = {Proceedings of The 8th Conference on Robot Learning (CoRL)},
  series    = {Proceedings of Machine Learning Research},
  volume    = {270},
  pages     = {1541--1566},
  year      = {2024},
  publisher = {PMLR},
  url       = {https://proceedings.mlr.press/v270/yuan25a.html}
}

@inproceedings{gloverpp,
  author    = {Ma, Teli and Zheng, Jia and Wang, Zifan and Gao, Ziyao and Zhou, Jiaming and Liang, Junwei},
  title     = {{GLOVER++}: Unleashing the Potential of Affordance Learning from Human Behaviors for Robotic Manipulation},
  booktitle = {Proceedings of The 9th Conference on Robot Learning (CoRL)},
  series    = {Proceedings of Machine Learning Research},
  volume    = {305},
  pages     = {3972--3994},
  year      = {2025},
  publisher = {PMLR},
  url       = {https://proceedings.mlr.press/v305/ma25b.html}
}

@inproceedings{dformer,
  title     = {{DFormerv2}: Geometry Self-Attention for {RGBD} Semantic Segmentation},
  author    = {Yin, Bo-Wen and Cao, Jiao-Long and Cheng, Ming-Ming and Hou, Qibin},
  booktitle = {Proceedings of the Computer Vision and Pattern Recognition Conference},
  pages     = {19345--19355},
  year      = {2025}
}

@inproceedings{opd,
  title     = {{OPD}: Single-view 3D Openable Part Detection},
  author    = {Jiang, Hanxiao and Mao, Yongsen and Savva, Manolis and Chang, Angel X.},
  booktitle = {European Conference on Computer Vision (ECCV)},
  year      = {2022},
}

@article{lin2025depthanything3,
  title   = {Depth Anything 3: Recovering the Visual Space from Any Views},
  author  = {Lin, Haotong and Chen, Sili and Liew, Junhao and Chen, Donny Y. and Li, Zhenyu and Shi, Guang and Feng, Jiashi and Kang, Bingyi},
  journal = {arXiv preprint arXiv:2511.10647},
  year    = {2025}
}

@inproceedings{ling2019curvegcn,
  title     = {Fast Interactive Object Annotation with {Curve-GCN}},
  author    = {Ling, Huan and Gao, Jun and Kar, Amlan and Chen, Wenzheng and Fidler, Sanja},
  booktitle = {Proceedings of the IEEE/CVF Conference on Computer Vision and Pattern Recognition (CVPR)},
  year      = {2019}
}

@article{krishna2017visualgenome,
  title   = {Visual Genome: Connecting Language and Vision Using Crowdsourced Dense Image Annotations},
  author  = {Krishna, Ranjay and Zhu, Yuke and Groth, Oliver and Johnson, Justin and Hata, Kenji and Kravitz, Joshua and Chen, Stephanie and Kalantidis, Yannis and Li, Li-Jia and Shamma, David A. and Bernstein, Michael S. and Fei-Fei, Li},
  journal = {International Journal of Computer Vision (IJCV)},
  volume  = {123},
  number  = {1},
  pages   = {32--73},
  year    = {2017}
}

@article{GarciaCastellanos2007,
  title = {Poles of inaccessibility: A calculation algorithm for the remotest places on earth},
  volume = {123},
  ISSN = {1751-665X},
  url = {http://dx.doi.org/10.1080/14702540801897809},
  DOI = {10.1080/14702540801897809},
  number = {3},
  journal = {Scottish Geographical Journal},
  publisher = {Informa UK Limited},
  author = {Garcia-Castellanos,  Daniel and Lombardo,  Umberto},
  year = {2007},
  month = sep,
  pages = {227-233}
}

@inproceedings{lai2023lisa,
  author    = {Lai, Xin and Tian, Zhuotao and Chen, Yukang and Li, Yanwei and Yuan, Yuhui and Liu, Shu and Jia, Jiaya},
  title     = {{LISA}: Reasoning Segmentation via Large Language Model},
  booktitle = {Proceedings of the IEEE/CVF Conference on Computer Vision and Pattern Recognition (CVPR)},
  pages     = {9579--9589},
  year      = {2024}
}

@article{vitra,
  title   = {{VITRA}: Scalable Vision-Language-Action Model Pretraining for Robotic Manipulation with Real-Life Human Activity Videos},
  author  = {{Microsoft VITRA Team}},
  journal = {arXiv preprint arXiv:2510.21571},
  year    = {2025}
}

@inproceedings{ego4d,
  title     = {{Ego4D}: Around the World in 3,000 Hours of Egocentric Video},
  author    = {Grauman, Kristen and Westbury, Andrew and Byrne, Eugene and Chavis, Zachary and Furnari, Antonino and Girdhar, Rohit and Hamburger, Jackson and Jiang, Hao and Liu, Miao and Liu, Xingyu and others},
  booktitle = {Proceedings of the IEEE/CVF Conference on Computer Vision and Pattern Recognition (CVPR)},
  pages     = {18995--19012},
  year      = {2022}
}

@article{epickitech,
  title   = {Rescaling Egocentric Vision: Collection, Pipeline and Challenges for {EPIC-KITCHENS-100}},
  author  = {Damen, Dima and Doughty, Hazel and Farinella, Giovanni Maria and Furnari, Antonino and Ma, Jian and Kazakos, Evangelos and Moltisanti, Davide and Munro, Jonathan and Perrett, Toby and Price, Will and Wray, Michael},
  journal = {International Journal of Computer Vision (IJCV)},
  volume  = {130},
  number  = {1},
  pages   = {33--55},
  year    = {2022}
}

@inproceedings{droid,
  title     = {{DROID}: A Large-Scale In-the-Wild Robot Manipulation Dataset},
  author    = {Khazatsky, Alexander and Pertsch, Karl and Nair, Suraj and Balakrishna, Ashwin and Dasari, Sudeep and Karamcheti, Siddharth and Nasiriany, Soroush and Srirama, Mohan Kumar and Chen, Lawrence Yunliang and Ellis, Kirsty and others},
  booktitle = {Proceedings of Robotics: Science and Systems (RSS)},
  year      = {2024},
  note      = {arXiv:2403.12945}
}

@article{rh20t,
  title   = {{RH20T}: A Comprehensive Robotic Dataset for Learning Diverse Skills in One-Shot},
  author  = {Fang, Hao-Shu and Fang, Hongjie and Tang, Zhenyu and Liu, Jirong and Wang, Junbo and Zhu, Haoyi and Lu, Cewu},
  journal = {arXiv preprint arXiv:2307.00595},
  year    = {2023}
}

@article{agiworld,
  title   = {{AgiBot World Colosseo}: A Large-scale Manipulation Platform for Scalable and Intelligent Embodied Systems},
  author  = {{AgiBot-World-Contributors} and Bu, Qingwen and Cai, Jisong and Chen, Li and Cui, Xiuqi and Ding, Yan and Feng, Siyuan and Gao, Shenyuan and He, Xindong and Huang, Xu and others},
  journal = {arXiv preprint arXiv:2503.06669},
  year    = {2025}
}

@article{robomind,
  title   = {{RoboMIND}: Benchmark on Multi-embodiment Intelligence Normative Data for Robot Manipulation},
  author  = {Wu, Kun and Hou, Chengkai and Liu, Jiaming and Che, Zhengping and Ju, Xiaozhu and Yang, Zhuqin and Li, Meng and Zhao, Yinuo and Xu, Zhiyuan and Yang, Guang and others},
  journal = {arXiv preprint arXiv:2412.13877},
  year    = {2024}
}

@article{robomind2,
  title   = {{RoboMIND 2.0}: A Multimodal, Bimanual Mobile Manipulation Dataset for Generalizable Embodied Intelligence},
  author  = {Hou, Chengkai and Wu, Kun and Liu, Jiaming and Che, Zhengping and Wu, Di and Liao, Fei and Li, Guangrun and He, Jingyang and Feng, Qiuxuan and Jin, Zhao and others},
  journal = {arXiv preprint arXiv:2512.24653},
  year    = {2025}
}

@article{calvin,
  title   = {{CALVIN}: A Benchmark for Language-Conditioned Policy Learning for Long-Horizon Robot Manipulation Tasks},
  author  = {Mees, Oier and Hermann, Lukas and Rosete-Beas, Erick and Burgard, Wolfram},
  journal = {IEEE Robotics and Automation Letters (RA-L)},
  volume  = {7},
  number  = {3},
  pages   = {7327--7334},
  year    = {2022}
}

@article{rlbench,
  title   = {{RLBench}: The Robot Learning Benchmark \& Learning Environment},
  author  = {James, Stephen and Ma, Zicong and Arrojo, David Rovick and Davison, Andrew J.},
  journal = {IEEE Robotics and Automation Letters (RA-L)},
  volume  = {5},
  number  = {2},
  pages   = {3019--3026},
  year    = {2020}
}

@article{lingbot,
  title   = {Masked Depth Modeling for Spatial Perception},
  author  = {Tan, Bin and Sun, Changjiang and Qin, Xiage and Adai, Hanat and Fu, Zelin and Zhou, Tianxiang and Zhang, Han and Xu, Yinghao and Zhu, Xing and Shen, Yujun and Xue, Nan},
  journal = {arXiv preprint arXiv:2601.17895},
  year    = {2026}
}

@inproceedings{sonata,
  title     = {Sonata: Self-Supervised Learning of Reliable Point Representations},
  author    = {Wu, Xiaoyang and DeTone, Daniel and Frost, Duncan and Shen, Tianwei and Xie, Chris and Yang, Nan and Engel, Jakob and Newcombe, Richard and Zhao, Hengshuang and Straub, Julian},
  booktitle = {Proceedings of the IEEE/CVF Conference on Computer Vision and Pattern Recognition (CVPR)},
  year      = {2025}
}

@article{sam3,
  title   = {{SAM 3}: Segment Anything with Concepts},
  author  = {{Meta AI Research}},
  journal = {arXiv preprint arXiv:2511.16719},
  year    = {2025}
}

@misc{qwen35blog,
    title = {Qwen3.5: Accelerating Productivity with Native Multimodal Agents},
    url = {https://qwen.ai/blog?id=qwen3.5},
    author = {Qwen Team},
    month = {February},
    year = {2026}
}

@inproceedings{delitzas2024scenefun3d,
  author    = {Delitzas, Alexandros and Takmaz, Ay{\c{c}}a and Tombari, Federico and Sumner, Robert and Pollefeys, Marc and Engelmann, Francis},
  title     = {{SceneFun3D}: Fine-Grained Functionality and Affordance Understanding in {3D} Scenes},
  booktitle = {Proceedings of the IEEE/CVF Conference on Computer Vision and Pattern Recognition (CVPR)},
  year      = {2024}
}

@inproceedings{nguyen2017object_crf,
  title={Object-Based Affordances Detection with Convolutional Neural Networks and Dense Conditional Random Fields},
  author={Nguyen, Anh and Kanoulas, Dimitrios and Caldwell, Darwin G and Tsagarakis, Nikos G},
  booktitle = {IEEE/RSJ International Conference on Intelligent Robots and Systems (IROS)},
  year={2017}
}

@inproceedings{do2018affordancenet,
  title={AffordanceNet: An End-to-End Deep Learning Approach for Object Affordance Detection},
  author={Do, Thanh-Toan and Nguyen, Anh and Reid, Ian},
  booktitle={International Conference on Robotics and Automation (ICRA)},
  year={2018}
}

@inproceedings{luo2022learning,
  title={Learning Affordance Grounding from Exocentric Images},
  author={Luo, Hongchen and Zhai, Wei and Zhang, Jing and Cao, Yang and Tao, Dacheng},
  booktitle={CVPR},
  year={2022}
}

@inProceedings{mo2021where2act,
    title={Where2Act: From Pixels to Actions for Articulated 3D Objects},
    author={Mo, Kaichun and Guibas, Leonidas and Mukadam, Mustafa and Gupta, Abhinav and Tulsiani, Shubham},
    year={2021},
    booktitle={International Conference on Computer Vision (ICCV)}
}

@inproceedings{wu2022vatmart,
title={{VAT}-Mart: Learning Visual Action Trajectory Proposals for Manipulating 3D {ART}iculated Objects},
author={Ruihai Wu and Yan Zhao and Kaichun Mo and Zizheng Guo and Yian Wang and Tianhao Wu and Qingnan Fan and Xuelin Chen and Leonidas Guibas and Hao Dong},
booktitle={International Conference on Learning Representations},
year={2022},
url={https://openreview.net/forum?id=iEx3PiooLy}
}

@inproceedings{li2023locate,
  title = {LOCATE: Localize and Transfer Object Parts for Weakly Supervised Affordance Grounding},
  author = {Li, Gen and Jampani, Varun and Sun, Deqing and Sevilla-Lara, Laura},
  booktitle={Proceedings of the IEEE/CVF Conference on Computer Vision and Pattern Recognition},
  year={2023}
}

@inproceedings{jang2024intra,
  author    = {Jang, Ji Ha and Seo, Hoigi and Chun, Se Young},
  title     = {{INTRA}: Interaction Relationship-aware Weakly Supervised Affordance Grounding},
  booktitle = {European Conference on Computer Vision (ECCV)},
  pages     = {18--34},
  year      = {2024},
  publisher = {Springer}
}

@inproceedings{xu2025partprior,
title={Weakly-Supervised Affordance Grounding Guided by Part-Level Semantic Priors},
author={Peiran Xu and Yadong MU},
booktitle={The Thirteenth International Conference on Learning Representations},
year={2025},
url={https://openreview.net/forum?id=0823rvTIhs}
}

@inproceedings{li2024precise,
      title     = {Learning Precise Affordances from Egocentric Videos for Robotic Manipulation},
      author    = {Li, Gen and Tsagkas, Nikolaos and Song, Jifei and Mon-Williams, Ruaridh and Vijayakumar, Sethu and Shao, Kun and Sevilla-Lara, Laura},
      booktitle = {Proceedings of the IEEE/CVF International Conference on Computer Vision},
      year      = {2025},
}

@InProceedings{heidinger2025_2handed,
  author    = {Heidinger, Marvin and Jauhri, Snehal and Prasad, Vignesh and Chalvatzaki, Georgia},
  title     = {2HandedAfforder: Learning Precise Actionable Bimanual Affordances from Human Videos},
  booktitle = {Proceedings of the IEEE/CVF International Conference on Computer Vision (ICCV)},
  month     = {October},
  year      = {2025},
  pages     = {14743-14753}
}

@inproceedings{huang2025bitalign,
  author       = {Yizhou Huang and
                  Fan Yang and
                  Guoliang Zhu and
                  Gen Li and
                  Hao Shi and
                  Yukun Zuo and
                  Wenrui Chen and
                  Zhiyong Li and
                  Kailun Yang},
  title        = {Resource-Efficient Affordance Grounding with Complementary Depth and
                  Semantic Prompts},
  booktitle    = {{IEEE/RSJ} International Conference on Intelligent Robots and Systems,
                  {IROS} 2025, Hangzhou, China, October 19-25, 2025},
  pages        = {7788--7795},
  publisher    = {{IEEE}},
  year         = {2025},
  url          = {https://doi.org/10.1109/IROS60139.2025.11245943},
  doi          = {10.1109/IROS60139.2025.11245943},
  bibsource    = {dblp computer science bibliography, https://dblp.org}
}

@misc{jiang2025affordancesam,
  author        = {Dengyang Jiang and
                   Zanyi Wang and
                   Hengzhuang Li and
                   Sizhe Dang and
                   Teli Ma and
                   Wei Wei and
                   Guang Dai and
                   Lei Zhang and
                   Mengmeng Wang},
  title         = {AffordanceSAM: Segment Anything Once More in Affordance Grounding},
  year          = {2025},
  eprint        = {2504.15650},
  archivePrefix = {arXiv},
  primaryClass  = {cs.CV},
  url           = {https://arxiv.org/abs/2504.15650}
}

@inproceedings{chen2024worldafford,
  author       = {Changmao Chen and
                  Yuren Cong and
                  Zhen Kan},
  title        = {Worldafford: Affordance Grounding Based on Natural Language Instructions},
  booktitle    = {36th {IEEE} International Conference on Tools with Artificial Intelligence,
                  {ICTAI} 2024, Herndon, VA, USA, October 28-30, 2024},
  pages        = {822--828},
  publisher    = {{IEEE}},
  year         = {2024},
  url          = {https://doi.org/10.1109/ICTAI62512.2024.00120},
  doi          = {10.1109/ICTAI62512.2024.00120},
  bibsource    = {dblp computer science bibliography, https://dblp.org}
}

@inproceedings{qian2024affordancellm,
  author       = {Shengyi Qian and
                  Weifeng Chen and
                  Min Bai and
                  Xiong Zhou and
                  Zhuowen Tu and
                  Li Erran Li},
  title        = {AffordanceLLM: Grounding Affordance from Vision Language Models},
  booktitle    = {{IEEE/CVF} Conference on Computer Vision and Pattern Recognition,
                  {CVPR} 2024 - Workshops, Seattle, WA, USA, June 17-18, 2024},
  pages        = {7587--7597},
  publisher    = {{IEEE}},
  year         = {2024},
  url          = {https://doi.org/10.1109/CVPRW63382.2024.00754},
  doi          = {10.1109/CVPRW63382.2024.00754},
  bibsource    = {dblp computer science bibliography, https://dblp.org}
}

@inproceedings{li2024manipllm,
  author       = {Xiaoqi Li and
                  Mingxu Zhang and
                  Yiran Geng and
                  Haoran Geng and
                  Yuxing Long and
                  Yan Shen and
                  Renrui Zhang and
                  Jiaming Liu and
                  Hao Dong},
  title        = {ManipLLM: Embodied Multimodal Large Language Model for Object-Centric
                  Robotic Manipulation},
  booktitle    = {{IEEE/CVF} Conference on Computer Vision and Pattern Recognition,
                  {CVPR} 2024, Seattle, WA, USA, June 16-22, 2024},
  pages        = {18061--18070},
  publisher    = {{IEEE}},
  year         = {2024},
  url          = {https://doi.org/10.1109/CVPR52733.2024.01710},
  doi          = {10.1109/CVPR52733.2024.01710},
  bibsource    = {dblp computer science bibliography, https://dblp.org}
}

@inproceedings{huang2024manipvqa,
  author       = {Siyuan Huang and
                  Iaroslav Ponomarenko and
                  Zhengkai Jiang and
                  Xiaoqi Li and
                  Xiaobin Hu and
                  Peng Gao and
                  Hongsheng Li and
                  Hao Dong},
  title        = {ManipVQA: Injecting Robotic Affordance and Physically Grounded Information
                  into Multi-Modal Large Language Models},
  booktitle    = {{IEEE/RSJ} International Conference on Intelligent Robots and Systems,
                  {IROS} 2024, Abu Dhabi, United Arab Emirates, October 14-18, 2024},
  pages        = {7580--7587},
  publisher    = {{IEEE}},
  year         = {2024},
  url          = {https://doi.org/10.1109/IROS58592.2024.10801993},
  doi          = {10.1109/IROS58592.2024.10801993},
  bibsource    = {dblp computer science bibliography, https://dblp.org}
}

@article{yoshida2024text,
  author       = {Tomoya Yoshida and
                  Shuhei Kurita and
                  Taichi Nishimura and
                  Shinsuke Mori},
  title        = {Text-driven affordance learning from egocentric vision},
  journal      = {Adv. Robotics},
  volume       = {39},
  number       = {16},
  pages        = {1041--1052},
  year         = {2025},
  url          = {https://doi.org/10.1080/01691864.2025.2535676},
  doi          = {10.1080/01691864.2025.2535676},
  bibsource    = {dblp computer science bibliography, https://dblp.org}
}

@inproceedings{fakp2025,
  author       = {Zhiyang Liu and
                  Ruiteng Zhao and
                  Lei Zhou and
                  Chengran Yuan and
                  Yuwei Wu and
                  Sheng Guo and
                  Zhengshen Zhang and
                  Chenchen Liu and
                  Marcelo H. Ang and
                  Francis E. H. Tay},
  title        = {3D Affordance Keypoint Detection for Robotic Manipulation},
  booktitle    = {{IEEE/RSJ} International Conference on Intelligent Robots and Systems,
                  {IROS} 2024, Abu Dhabi, United Arab Emirates, October 14-18, 2024},
  pages        = {7528--7534},
  publisher    = {{IEEE}},
  year         = {2024},
  url          = {https://doi.org/10.1109/IROS58592.2024.10801792},
  doi          = {10.1109/IROS58592.2024.10801792},
  bibsource    = {dblp computer science bibliography, https://dblp.org}
}

@inproceedings{ju2024robo_abc,
  title={Robo-abc: Affordance generalization beyond categories via semantic correspondence for robot manipulation},
  author={Ju, Yuanchen and Hu, Kaizhe and Zhang, Guowei and Zhang, Gu and Jiang, Mingrun and Xu, Huazhe},
  booktitle={European Conference on Computer Vision},
  pages={222--239},
  year={2024},
  organization={Springer}
}

@inproceedings{kuang2024ram,
  author    = {Kuang, Yuxuan and Ye, Junjie and Geng, Haoran and Mao, Jiageng and Deng, Congyue and Guibas, Leonidas and Wang, He and Wang, Yue},
  title     = {{RAM}: Retrieval-Based Affordance Transfer for Generalizable Zero-Shot Robotic Manipulation},
  booktitle = {Proceedings of The 8th Conference on Robot Learning (CoRL)},
  series    = {Proceedings of Machine Learning Research},
  volume    = {270},
  pages     = {547--565},
  year      = {2024},
  publisher = {PMLR},
  url       = {https://proceedings.mlr.press/v270/kuang25a.html}
}

@inproceedings{deng20213daffordnet,
  title={3D AffordanceNet: A Benchmark for Visual Object Affordance Understanding},
  author={Deng, Shengheng and Xu, Xun and Wu, Chaozheng and Chen, Ke and Jia, Kui},
  booktitle={Proceedings of the IEEE Conference on Computer Vision and Pattern Recognition},
  year={2021}
}

@InProceedings{yang2023iagnet,
    author    = {Yang, Yuhang and Zhai, Wei and Luo, Hongchen and Cao, Yang and Luo, Jiebo and Zha, Zheng-Jun},
    title     = {Grounding 3D Object Affordance from 2D Interactions in Images},
    booktitle = {Proceedings of the IEEE/CVF International Conference on Computer Vision (ICCV)},
    month     = {October},
    year      = {2023},
    pages     = {10905-10915}
}

@inproceedings{nguyen2023openad,
  author    = {Nguyen, Toan and Vu, Minh Nhat and Vuong, An and Nguyen, Dzung and Vo, Thieu and Le, Ngan and Nguyen, Anh},
  title     = {Open-Vocabulary Affordance Detection in {3D} Point Clouds},
  booktitle = {IEEE/RSJ International Conference on Intelligent Robots and Systems (IROS)},
  year      = {2023}
}

@inproceedings{chu2025_3daffordllm,
  author       = {Hengshuo Chu and
                  Xiang Deng and
                  Qi Lv and
                  Xiaoyang Chen and
                  Yinchuan Li and
                  Jianye Hao and
                  Liqiang Nie},
  title        = {3D-AffordanceLLM: Harnessing Large Language Models for Open-Vocabulary
                  Affordance Detection in 3D Worlds},
  booktitle    = {The Thirteenth International Conference on Learning Representations,
                  {ICLR} 2025, Singapore, April 24-28, 2025},
  publisher    = {OpenReview.net},
  year         = {2025},
  url          = {https://openreview.net/forum?id=GThTiuXgDC},
  bibsource    = {dblp computer science bibliography, https://dblp.org}
}

@inproceedings{yu2025seqafford,
  author    = {Yu, Chunlin and Wang, Hanqing and Shi, Ye and Luo, Haoyang and Yang, Sibei and Yu, Jingyi and Wang, Jingya},
  title     = {{SeqAfford}: Sequential {3D} Affordance Reasoning via Multimodal Large Language Model},
  booktitle = {Proceedings of the IEEE/CVF Conference on Computer Vision and Pattern Recognition (CVPR)},
  pages     = {1691--1701},
  year      = {2025}
}

@inproceedings{shao2025great,
  author    = {Shao, Yawen and Zhai, Wei and Yang, Yuhang and Luo, Hongchen and Cao, Yang and Zha, Zheng-Jun},
  title     = {{GREAT}: Geometry-Intention Collaborative Inference for Open-Vocabulary {3D} Object Affordance Grounding},
  booktitle = {Proceedings of the IEEE/CVF Conference on Computer Vision and Pattern Recognition (CVPR)},
  pages     = {17326--17336},
  year      = {2025}
}

@InProceedings{lu2025geal,
    author    = {Lu, Dongyue and Kong, Lingdong and Huang, Tianxin and Lee, Gim Hee},
    title     = {GEAL: Generalizable 3D Affordance Learning with Cross-Modal Consistency},
    booktitle = {Proceedings of the IEEE/CVF Conference on Computer Vision and Pattern Recognition (CVPR)},
    month     = {June},
    year      = {2025},
    pages     = {1680-1690}
}

@inproceedings{zhu2025lmaffordance3d,
  author       = {He Zhu and
                  Quyu Kong and
                  Kechun Xu and
                  Xunlong Xia and
                  Bing Deng and
                  Jieping Ye and
                  Rong Xiong and
                  Yue Wang},
  title        = {Grounding 3D Object Affordance with Language Instructions, Visual
                  Observations and Interactions},
  booktitle    = {{IEEE/CVF} Conference on Computer Vision and Pattern Recognition,
                  {CVPR} 2025, Nashville, TN, USA, June 11-15, 2025},
  pages        = {17337--17346},
  publisher    = {Computer Vision Foundation / {IEEE}},
  year         = {2025},
  url          = {https://openaccess.thecvf.com/content/CVPR2025/html/Zhu\_Grounding\_3D\_Object\_Affordance\_with\_Language\_Instructions\_Visual\_Observations\_and\_CVPR\_2025\_paper.html},
  doi          = {10.1109/CVPR52734.2025.01616},
  bibsource    = {dblp computer science bibliography, https://dblp.org}
}

@inproceedings{o3afford,
  author    = {Tian, Tongxuan and Kang, Xuhui and Kuo, Yen-Ling},
  title     = {O3Afford: One-Shot 3D Object-to-Object Affordance Grounding for Generalizable Robotic Manipulation},
  journal   = {Conference on Robot Learning (CoRL)},
  year      = {2025},
}

@article{semantic_funcmap,
  author       = {Xiaoxiang Dong and
                  Weiming Zhi},
  title        = {Affordance Transfer Across Object Instances via Semantically Anchored
                  Functional Map},
  journal      = {CoRR},
  volume       = {abs/2602.14874},
  year         = {2026},
  url          = {https://doi.org/10.48550/arXiv.2602.14874},
  doi          = {10.48550/ARXIV.2602.14874},
  eprinttype   = {arXiv},
  eprint       = {2602.14874},
  bibsource    = {dblp computer science bibliography, https://dblp.org}
}

@misc{videoafford,
  author        = {Hanqing Wang and
                   Mingyu Liu and
                   Xiaoyu Chen and
                   Chengwei Ma and
                   Yiming Zhong and
                   Wenti Yin and
                   Yuhao Liu and
                   Zhiqing Cui and
                   Jiahao Yuan and
                   Lu Dai and
                   Zhiyuan Ma and
                   Hui Xiong},
  title         = {VideoAfford: Grounding 3D Affordance from Human-Object-Interaction Videos via Multimodal Large Language Model},
  year          = {2026},
  eprint        = {2602.09638},
  archivePrefix = {arXiv},
  primaryClass  = {cs.CV},
  url           = {https://arxiv.org/abs/2602.09638}
}

@misc{vagnet,
  author        = {Aihua Mao and
                   Kaihang Huang and
                   Yong-Jin Liu and
                   Chee Seng Chan and
                   Ying He},
  title         = {VAGNet: Grounding 3D Affordance from Human-Object Interactions in Videos},
  year          = {2026},
  eprint        = {2602.20608},
  archivePrefix = {arXiv},
  primaryClass  = {cs.CV},
  url           = {https://arxiv.org/abs/2602.20608}
}

@inproceedings{affordbot,
  author    = {Wang, Xinyi and Yang, Xun and Xu, Yanlong and Wu, Yuchen and Li, Zhen and Zhao, Na},
  title     = {{AffordBot}: {3D} Fine-grained Embodied Reasoning via Multimodal Large Language Models},
  booktitle = {Advances in Neural Information Processing Systems (NeurIPS)},
  year      = {2025}
}

@inproceedings{corona2020ganhand,
  author    = {Corona, Enric and Pumarola, Albert and Alenya, Guillem and Moreno-Noguer, Francesc and Rogez, Gr{\'e}gory},
  title     = {{GanHand}: Predicting Human Grasp Affordances in Multi-Object Scenes},
  booktitle = {Proceedings of the IEEE/CVF Conference on Computer Vision and Pattern Recognition (CVPR)},
  pages     = {5031--5041},
  year      = {2020}
}

@inproceedings{affordgrasp,
  author       = {Yingbo Tang and
                  Shuaike Zhang and
                  Xiaoshuai Hao and
                  Pengwei Wang and
                  Jianlong Wu and
                  Zhongyuan Wang and
                  Shanghang Zhang},
  title        = {AffordGrasp: In-Context Affordance Reasoning for Open-Vocabulary Task-Oriented
                  Grasping in Clutter},
  booktitle    = {{IEEE/RSJ} International Conference on Intelligent Robots and Systems,
                  {IROS} 2025, Hangzhou, China, October 19-25, 2025},
  pages        = {9433--9439},
  publisher    = {{IEEE}},
  year         = {2025},
  url          = {https://doi.org/10.1109/IROS60139.2025.11245995},
  doi          = {10.1109/IROS60139.2025.11245995},
  bibsource    = {dblp computer science bibliography, https://dblp.org}
}

@inproceedings{afford_dex_grasp,
  author    = {Wei, Yi-Lin and Lin, Mu and Lin, Yuhao and Jiang, Jian-Jian and Wu, Xiao-Ming and Zeng, Ling-An and Zheng, Wei-Shi},
  title     = {{AffordDexGrasp}: Open-set Language-guided Dexterous Grasp with Generalizable-Instructive Affordance},
  booktitle = {Proceedings of the IEEE/CVF International Conference on Computer Vision (ICCV)},
  year      = {2025}
}

@misc{fsag,
  author        = {Yifan Han and
                   Yichuan Peng and
                   Pengfei Yi and
                   Junyan Li and
                   Hanqing Wang and
                   Gaojing Zhang and
                   Qi Peng Liu and
                   Wenzhao Lian},
  title         = {FSAG: Enhancing Human-to-Dexterous-Hand Finger-Specific Affordance Grounding via Diffusion Models},
  year          = {2026},
  eprint        = {2601.08246},
  archivePrefix = {arXiv},
  primaryClass  = {cs.RO},
  url           = {https://arxiv.org/abs/2601.08246}
}

@article{mllm_afford_grasp,
  author       = {Zhou Zhao and
                  Jie Gao and
                  Dongyuan Zheng},
  title        = {Affordance-Guided Robotic Grasping via Multimodal Large Language Model
                  Reasoning},
  journal      = {{IEEE} Trans Autom. Sci. Eng.},
  volume       = {23},
  pages        = {4088--4100},
  year         = {2026},
  url          = {https://doi.org/10.1109/TASE.2026.3651854},
  doi          = {10.1109/TASE.2026.3651854},
  bibsource    = {dblp computer science bibliography, https://dblp.org}
}

@inproceedings{yu2024uniaff,
  author       = {Qiaojun Yu and
                  Siyuan Huang and
                  Xibin Yuan and
                  Zhengkai Jiang and
                  Ce Hao and
                  Xin Li and
                  Haonan Chang and
                  Junbo Wang and
                  Liu Liu and
                  Hongsheng Li and
                  Peng Gao and
                  Cewu Lu},
  title        = {UniAff: {A} Unified Representation of Affordances for Tool Usage and
                  Articulation with Vision-Language Models},
  booktitle    = {{IEEE} International Conference on Robotics and Automation, {ICRA}
                  2025, Atlanta, GA, USA, May 19-23, 2025},
  pages        = {8980--8987},
  publisher    = {{IEEE}},
  year         = {2025},
  url          = {https://doi.org/10.1109/ICRA55743.2025.11127736},
  doi          = {10.1109/ICRA55743.2025.11127736},
  bibsource    = {dblp computer science bibliography, https://dblp.org}
}

@inproceedings{opdmulti,
  author       = {Xiaohao Sun and
                  Hanxiao Jiang and
                  Manolis Savva and
                  Angel X. Chang},
  title        = {{OPDMulti}: Openable Part Detection for Multiple Objects},
  booktitle    = {International Conference on 3D Vision, 3DV 2024, Davos, Switzerland,
                  March 18-21, 2024},
  pages        = {169--178},
  publisher    = {{IEEE}},
  year         = {2024},
  url          = {https://doi.org/10.1109/3DV62453.2024.00100},
  doi          = {10.1109/3DV62453.2024.00100}
}

@inproceedings{mopd,
  author       = {Siqi Li and
                  Xiaoxue Chen and
                  Haoyu Cheng and
                  Guyue Zhou and
                  Hao Zhao and
                  Guanzhong Tian},
  editor       = {Minsu Cho and
                  Ivan Laptev and
                  Du Tran and
                  Angela Yao and
                  Hongbin Zha},
  title        = {Locate {n'} {Rotate}: {Two-stage} Openable Part Detection with Geometric Foundation Model Priors},
  booktitle    = {Computer Vision - {ACCV} 2024 - 17th Asian Conference on Computer
                  Vision, Hanoi, Vietnam, December 8-12, 2024, Proceedings, Part {VII}},
  series       = {Lecture Notes in Computer Science},
  pages        = {716--732},
  publisher    = {Springer},
  year         = {2024},
  url          = {https://doi.org/10.1007/978-981-96-0963-5_6},
  doi          = {10.1007/978-981-96-0963-5\_6}
}

@inproceedings{yoshida2025_6dof,
  author    = {Yoshida, Tomoya and Kurita, Shuhei and Nishimura, Taichi and Mori, Shinsuke},
  title     = {Generating {6DoF} Object Manipulation Trajectories from Action Description in Egocentric Vision},
  booktitle = {Proceedings of the IEEE/CVF Conference on Computer Vision and Pattern Recognition (CVPR)},
  month     = {June},
  year      = {2025},
  pages     = {17370--17382}
}

@inproceedings{flowbot3d,
    author    = {Ben Eisner and Harry Zhang and David Held},
    title     = {{FlowBot3D}: Learning {3D} Articulation Flow to Manipulate Articulated Objects},
    booktitle = {Proceedings of Robotics: Science and Systems},
    year      = {2022},
    address   = {New York City, NY, USA},
    month     = {June},
    doi       = {10.15607/RSS.2022.XVIII.018}
}

@inproceedings{flowbotpp,
  author       = {Harry Zhang and
                  Ben Eisner and
                  David Held},
  editor       = {Jie Tan and
                  Marc Toussaint and
                  Kourosh Darvish},
  title        = {{FlowBot++}: Learning Generalized Articulated Objects Manipulation via
                  Articulation Projection},
  booktitle    = {Conference on Robot Learning, CoRL 2023, 6-9 November 2023, Atlanta,
                  GA, {USA}},
  series       = {Proceedings of Machine Learning Research},
  pages        = {1222--1241},
  publisher    = {{PMLR}},
  year         = {2023},
  url          = {https://proceedings.mlr.press/v229/zhang23c.html}
}

@inproceedings{track2act,
  author    = {Bharadhwaj, Homanga and Mottaghi, Roozbeh and Gupta, Abhinav and Tulsiani, Shubham},
  title     = {{Track2Act}: Predicting Point Tracks from Internet Videos enables Generalizable Robot Manipulation},
  booktitle = {European Conference on Computer Vision (ECCV)},
  year      = {2024},
  publisher = {Springer}
}

@inproceedings{toolflownet,
    title     = {{ToolFlowNet}: Robotic Manipulation with Tools via Predicting Tool Flow from Point Clouds},
    author    = {Seita, Daniel and Wang, Yufei and Shetty, Sarthak and Li, Edward and Erickson, Zackory and Held, David},
    booktitle = {Conference on Robot Learning (CoRL)},
    year      = {2022}
}

@inproceedings{afforddp,
  title     = {{AffordDP}: Generalizable Diffusion Policy with Transferable Affordance},
  author    = {Wu, Shijie and Zhu, Yihang and Huang, Yunao and Zhu, Kaizhen and Gu, Jiayuan and Yu, Jingyi and Shi, Ye and Wang, Jingya},
  booktitle = {Proceedings of the Computer Vision and Pattern Recognition Conference (CVPR)},
  pages     = {6971--6980},
  year      = {2025}
}

@article{afford_flow_matching,
  author       = {Fan Zhang and
                  Michael Gienger},
  title        = {Affordance-based Robot Manipulation with Flow Matching},
  journal      = {CoRR},
  volume       = {abs/2409.01083},
  year         = {2024},
  url          = {https://doi.org/10.48550/arXiv.2409.01083},
  doi          = {10.48550/ARXIV.2409.01083},
  eprinttype   = {arXiv},
  eprint       = {2409.01083}
}

@article{rt_affordance,
  author       = {Soroush Nasiriany and
                  Sean Kirmani and
                  Tianli Ding and
                  Laura Smith and
                  Yuke Zhu and
                  Danny Driess and
                  Dorsa Sadigh and
                  Ted Xiao},
  title        = {{RT-Affordance}: Affordances are Versatile Intermediate Representations
                  for Robot Manipulation},
  journal      = {CoRR},
  volume       = {abs/2411.02704},
  year         = {2024},
  url          = {https://doi.org/10.48550/arXiv.2411.02704},
  doi          = {10.48550/ARXIV.2411.02704},
  eprinttype   = {arXiv},
  eprint       = {2411.02704}
}

@article{anchordp3,
  author       = {Ziyan Zhao and
                  Ke Fan and
                  He{-}Yang Xu and
                  Ning Qiao and
                  Bo Peng and
                  Wenlong Gao and
                  Dongjiang Li and
                  Hui Shen},
  title        = {{AnchorDP3}: {3D} Affordance Guided Sparse Diffusion Policy for Robotic
                  Manipulation},
  journal      = {CoRR},
  volume       = {abs/2506.19269},
  year         = {2025},
  url          = {https://doi.org/10.48550/arXiv.2506.19269},
  doi          = {10.48550/ARXIV.2506.19269},
  eprinttype   = {arXiv},
  eprint       = {2506.19269}
}

@inproceedings{g3flow,
    author    = {Chen, Tianxing and Mu, Yao and Liang, Zhixuan and Chen, Zanxin and Peng, Shijia and Chen, Qiangyu and Xu, Mingkun and Hu, Ruizhen and Zhang, Hongyuan and Li, Xuelong and Luo, Ping},
    title     = {{G3Flow}: Generative {3D} Semantic Flow for Pose-aware and Generalizable Object Manipulation},
    booktitle = {Proceedings of the IEEE/CVF Conference on Computer Vision and Pattern Recognition (CVPR)},
    month     = {June},
    year      = {2025},
    pages     = {1735--1744}
}

@inproceedings{myers2015umd,
  author    = {Austin Myers and Ching L. Teo and Cornelia Ferm\"uller and Yiannis Aloimonos},
  title     = {Affordance Detection of Tool Parts from Geometric Features},
  booktitle = {ICRA},
  year      = {2015}
}

@inproceedings{roboafford,
  title     = {RoboAfford: A Dataset and Benchmark for Enhancing Object and Spatial Affordance Learning in Robot Manipulation},
  author    = {Tang, Yingbo and Zhang, Lingfeng and Zhang, Shuyi and Zhao, Yinuo and Hao, Xiaoshuai},
  booktitle = {Proceedings of the 33rd ACM International Conference on Multimedia},
  pages     = {12706--12713},
  year      = {2025}
}

@article{roboafford_pp,
  title   = {RoboAfford++: A Generative AI-Enhanced Dataset for Multimodal Affordance Learning in Robotic Manipulation and Navigation},
  author  = {Hao, Xiaoshuai and Tang, Yingbo and Zhang, Lingfeng and Ma, Yanbiao and Diao, Yunfeng and Jia, Ziyu and Ding, Wenbo and Ye, Hangjun and Chen, Long},
  journal = {arXiv preprint arXiv:2511.12436},
  year    = {2025}
}

@article{egoscaler,
  author       = {Tomoya Yoshida and
                  Shuhei Kurita and
                  Taichi Nishimura and
                  Shinsuke Mori},
  title        = {Developing Vision-Language-Action Model from Egocentric Videos},
  journal      = {CoRR},
  volume       = {abs/2509.21986},
  year         = {2025},
  url          = {https://doi.org/10.48550/arXiv.2509.21986},
  doi          = {10.48550/ARXIV.2509.21986},
  eprinttype   = {arXiv},
  eprint       = {2509.21986}
}

@inproceedings{hot3d,
  author    = {Banerjee, Prithviraj and Shkodrani, Sindi and Moulon, Pierre and Hampali, Shreyas and Han, Shangchen and Zhang, Fan and Zhang, Linguang and Fountain, Jade and Miller, Edward and Basol, Selen and Newcombe, Richard and Wang, Robert and Engel, Jakob Julian and Hodan, Tomas},
  title     = {{HOT3D}: Hand and Object Tracking in {3D} from Egocentric Multi-View Videos},
  booktitle = {Proceedings of the IEEE/CVF Conference on Computer Vision and Pattern Recognition (CVPR)},
  pages     = {7061--7071},
  year      = {2025}
}

\newpage
\appendix

\section*{Technical Appendices and Supplementary Material}
\label{sec:appendix}

This appendix provides supplementary material supporting the claims in the
main paper.
\textbf{App.~\ref{sec:app_limit}} specifies the limitations and failed case analysis of our method, and social responsibility of our work.
\textbf{App.~\ref{sec:app_data}} describes how heterogeneous sources are
converted into our unified affordance-data schema, covering
source-specific preprocessing, SAM3 query generation, curve fitting, \ourdataset galleries, and the HOVA-500K
conversion used for segmentation training.
\textbf{App.~\ref{sec:app_eval}} specifies more details of our experiment, with additional qualitative results, and more robotic demo examples.

\section{Limitations, Failed Case Analysis and Social Responsibility}
\label{sec:app_limit}

\begin{figure}[h]
    \centering
    \includegraphics[width=0.6\linewidth]{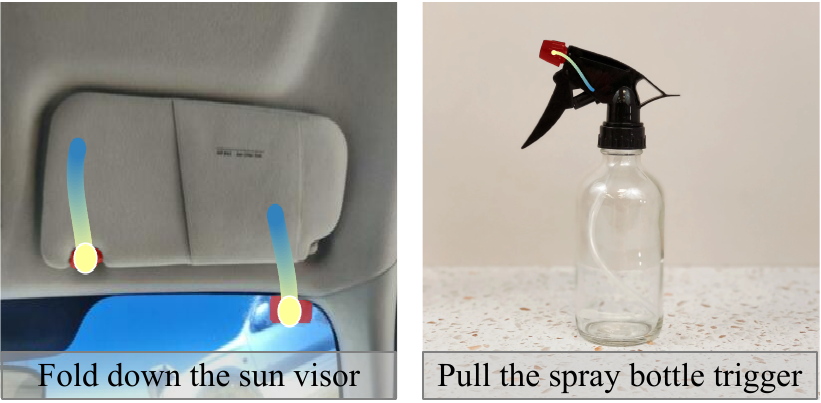}
    \caption{Example of Failed Cases. As there is nothing similar to Spray Bottles or Sun Visors in the training dataset, our model struggles in such cases.}
    \label{fig:app_limit_examples}
\end{figure}

\paragraph{Limitations and Failure Cases.}
Though our model has the capability to adapt to open-world images based on our motion dataset, this adaptation is still limited if such motion is completely novel.
We show two examples in Figure~\ref{fig:app_limit_examples}. There are no similar objects for model to relate such action to. 
As a result, it cannot accurately predict how the motion would go regarding these tasks. A valid next step would be to create an even larger-scale dataset, truly on the scale of millions, and to be truly open-world.
\paragraph{Social Impact and Responsibility.}
\ourmodel may support more interpretable robot manipulation by predicting inspectable contact regions and post-contact motion, and its data pipeline may help standardize affordance supervision across heterogeneous sources.
However, it is not a standalone policy or safety system: physical deployment should include human supervision, collision checking, force and workspace limits, emergency stops, and task-specific validation, especially near people, fragile objects, or hazardous materials.
Because improved manipulation models can be misused when paired with capable robots, we will release the work with clear intended-use guidance, documented failure modes, and source-dataset, license, privacy, and redistribution constraints; downstream users should evaluate bias and generalization in their own environments rather than assuming benchmark performance transfers uniformly.

\section{Dataset Pipeline Details}
\label{sec:app_data}
\label{sec:appendix-data-statistics}

We turn heterogeneous demonstrations from robot teleoperation,
human egocentric video, simulation data, and real-world scans into a unified
affordance dataset through a two-phase pipeline. First, a dataset-specific \emph{preprocess} module handles raw-format differences and exports a shared per-interval schema. Then, a dataset-agnostic \emph{mainprocess} consumes this schema to generate training data: a SAM3 task query, a tracked object mask, depth, a 3D object trajectory, and fitted \curverep parameters. This design keeps format handling separate from affordance label extraction, so adding a new data source only requires writing a new preprocess adapter. Per-source statistics are reported in Table~\ref{tab:appendix_data_stats}.

\begin{table}[t]
\centering
\caption{Per-source dataset statistics. \emph{Episodes} refer to the top-level recording units in each source; \emph{intervals} are action-segmented sub-clips of an episode; \emph{views} are per-camera observations of an interval; \emph{fitted curves} are views with a successfully fit \curverep.}
\label{tab:appendix_data_stats}
\resizebox{\columnwidth}{!}{%
\begin{tabular}{l l rrrr}
\toprule
Source & Modality & Episodes & Intervals & Views & Fitted curves \\
\midrule
agibot & robot teleop & 17,124 & 25,395 & 25,395 & 1,493 \\
droid & robot teleop & 47,508 & 66,212 & 132,424 & 7,381 \\
rh20t & robot teleop & 3,588 & 4,512 & 34,163 & 4,127 \\
robomind & robot teleop & 14,695 & 22,642 & 48,826 & 17,198 \\
robomind2 & robot teleop & 6,738 & 8,918 & 28,663 & 4,965 \\
\midrule
hoi4d & human egocentric & 1,020 & 2,165 & 2,165 & 1,974 \\
vitra & human egocentric & 3,024 & 1,098,944 & 1,098,944 & 129,433 \\
\midrule
calvin & simulation & 181 & 6,649 & 6,649 & 4,002 \\
rlbench & simulation & 268 & 276 & 1,104 & 69 \\
\midrule
scenefun3d & real scan & 585 & 7,027 & 148,603 & 52,692 \\
\midrule
\textbf{Total} & & \textbf{94,731} & \textbf{1,242,740} & \textbf{1,526,936} & \textbf{223,334} \\
\bottomrule
\end{tabular}%
}
\end{table}

\subsection{Cross-Dataset Preprocess}
\label{sec:appendix-data-preprocessing}
\label{sec:appendix-preprocess}

Our data preprocess pipeline is built around source-specific adapters that share
the same processing logic and structure. Every adapter writes the
same interval-level schema: observation/contact frames, RGB-D, task
language, camera calibration, and the video span. What differs is how
these fields are recovered from each raw dataset. Below we summarize the
dataset-specific handling.

\paragraph{AgiBot.}
AgiBotWorld-Beta provides action-interval annotations, and we use each
annotated action as one interval. We load the paired head RGB/depth
frames for the interval and use the calibrated non-fisheye head camera.

\paragraph{DROID.}
DROID provides ZED stereo recordings together with synchronized robot and
camera metadata. We decode the left RGB-D stream from each external
stereo camera. Calibration comes from the official patched files when
available; otherwise we fall back to the original HDF5 metadata. We attach language from the patched files and discard runs marked as failures.

\paragraph{RH20T.}
For RH20T, manipulation spans are not annotated directly. We infer them
from the gripper-command trace: a new interval starts when the gripper
begins to close and ends when it opens again. The interval end is
extended slightly after release to preserve post-contact motion. We
export only static camera views and remove intervals that are too short
to provide a meaningful motion trajectory.

\paragraph{RoboMIND.}
RoboMIND-v1 mixes several robot embodiments, so the main preprocessing
issue is to make their camera layouts and annotations consistent. We
identify the static camera views for each embodiment and discard
arm-mounted views. RGB-D streams are decoded and depth is normalized to a
common metric scale. Manipulation intervals come from the per-step
language annotations released with RoboMIND, which provide frame
boundaries for each step.

\paragraph{RoboMIND-v2.}
RoboMIND-v2 combines Tien Kung, Franka, UR5, and Ark recordings, and the
main challenge is that their gripper signals and usable camera views are
not encoded in the same way. We first identify the robot family for each
episode and keep only the camera views that can be used reliably in our
pipeline. Manipulation intervals are then recovered from the family-specific gripper signal; Ark requires an additional range check because
its recordings use two different gripper-value encodings. We fall back to
the whole episode when no valid gripper interval is found, and we remove
non-rigid tasks such as cloth, folding, and rope. Note that this dataset is completely excluded from training set and only used as testing set.

\paragraph{HOI4D.}
HOI4D differs from robot sources because the camera is moving, but the
dataset also provides precise geometric annotations: action event
markers, camera extrinsics, 3D object assets, object masks, and
object pose trajectories. Because these annotations already determine
both the temporal span and the 3D object motion, we handle HOI4D through
a custom path rather than the robot adapter interface. We use event
boundaries such as \emph{Reachout}, \emph{Grasp}, and \emph{Pickup} to
form the manipulation interval, and recover the motion label directly
from the recorded object pose trajectory instead of running depth-based
mask tracking. Since the provided masks are object-level, we still run
SAM to obtain the part-level mask used by our
affordance supervision.

\paragraph{VITRA.}
VITRA is our main source of human manipulation clips. One main challenge
for egocentric human videos is noisy camera motion, and VITRA provides
per-frame SLAM camera intrinsics and extrinsics for EPIC-KITCHENS and
Ego4D clips. We therefore use these camera poses to make the trajectories
geometrically usable. When predicted depth is used, a tracked mask point
is first back-projected in the camera frame where it is observed; the
per-frame SLAM poses then transform this 3D point through the world frame
into the observation-frame camera. Since VITRA stores annotations
separately from the source videos, we first resolve each annotated frame
index back to the corresponding EPIC-KITCHENS or Ego4D frame. We use
contact-index files to reject clips without real hand-object contact, and
express all projected quantities in the observation-frame camera
coordinates.

\paragraph{Calvin.}
Calvin is a simulated robot manipulation dataset with task language,
RGB-D observations, and robot-state traces. We use only the static scene
camera, and discard gripper-mounted and tactile views because they do not
provide a stable view of the scene. Calvin's language annotations mark
task-level windows in a continuous rollout, but they are not contact
markers. We use them as semantic anchors, then use the gripper-action
trace to refine the temporal span: an annotation is kept only when a
nearby gripper-closing segment is found, and the final interval covers
both the language window and the matched interaction motion. We use the
static RGB-D observation for 3D back-projection, but do not use
simulator-only object variables, such as object poses, drawer states, or
button states, as supervision.

\paragraph{RLBench.}
RLBench is a simulated robot manipulation dataset with task language,
RGB-D observations, robot poses, and a binary gripper-open signal. We
keep the static scene cameras and exclude the wrist camera because it
moves with the arm. RLBench provides task language at the episode level
rather than per-step temporal spans. We therefore recover intervals from
the gripper signal when possible: closed-gripper segments become
manipulation intervals, with the observation frame shifted earlier to
include the approach. Simulator depth is converted to metric depth for
3D back-projection, but we do not use privileged simulator outputs, such
as ground-truth masks or object states, to generate affordance labels.

\paragraph{SceneFun3D.}
SceneFun3D provides posed ARKit RGB-D views of scanned rooms. Each task
is tied to an annotated 3D affordance region and a motion primitive.
Unlike the robot or human action interval, there is no object movement in the dataset. 
We therefore bypass the tracking, projection, and curve fitting stages in our data pipeline for SceneFun3D.
We sampled the frames at 10 FPS in the dataset videos. 
We keep steps 2 and 3 (green block in Fig.~\ref{fig:data_pipeline}) to retrieve masks of the object of interest,
and keep the frames where that object's mask is near the starting point of the motion annotation and the projected object 3D annotation.
We then directly convert the motion trajectory to our \curverep representation, taking a radius of 90 degrees for circular motion and a fixed 0.3 m length for linear motion.

We use a single set of unit and frame conventions throughout: depth
maps in millimeters, 3D positions in meters, and
rigid transforms expressed as $\mathbf{T}_{\text{base}\to\text{cam}}$.
Every adapter is unit-tested against this contract before integration.

\subsection{Mainprocess}
\label{sec:appendix-mainprocess}
\label{sec:appendix-tracking}
\paragraph{Query Generation via vLLM Qwen3.5.}
\label{sec:appendix-querygen}

The goal of query generation is to produce a text query that lets
SAM3~\citep{sam3} segments the affordance mask for each task interval.
This mask should cover the object part, visible in the observation
image, where contact should be made to execute the task.
Since SAM3 can be driven by text prompts,
we first convert the interval's high-level instruction $y$ (e.g.,
``\texttt{Open the cabinet door.}'') and visual evidence
$\{I_{\mathrm{obs}}, I_{\mathrm{contact}}\}$ into a short,
open-vocabulary segmentation phrase $q$. The phrase is required to name
the \emph{minimal manipulable target part}, such as a handle, knob,
button, latch, or lid edge, rather than restating the action or naming
the whole object when a smaller contact part is visible. The observation
frame determines how the part should be described in the target image,
while the contact frame helps disambiguate which part is actually used.
The downstream SAM3 video predictor is text-prompted with $q$, so the precision of $q$ directly affects both
the affordance mask and the recovered $3$D motion.

We use Qwen3.5-35B-A3B-FP8~\citep{qwen35blog} to generate the SAM3 task query. The model returns a JSON object with a brief rationale and
the final \texttt{sam3\_prompt}. The full system and user prompts
are shown below.

\begingroup\small
\begin{verbatim}
[SYSTEM]
You are an expert at converting robotics task language + visual
evidence into a compact open-vocabulary segmentation query for SAM3.

Project context — Affordance Understanding:
This data is used to train an affordance prediction model. Given an
RGB image and a task-level instruction (e.g., "Open the cabinet
door"), the model must predict (1) the **affordance region** — the
spatial area on the object where physical contact should occur, and
(2) the **post-contact motion trajectory**. The task description
intentionally specifies *what to do*, NOT *where to touch*; the model
must learn the functional mapping from task intent to contact region.
Therefore, only data samples with clear, physically grounded
manipulation targets and meaningful task-level semantics are valuable
for training.

Your goal is to produce a short noun phrase that uniquely identifies
the **smallest manipulable target part** (e.g., "top-left drawer
handle", "right knob", "front latch", "left hinge", "power button")
that the robot will operate in the observation image.
You must be extremely concise and precise. You must not describe
actions; only describe the target object/part to segment. Prefer
concrete part names + discriminative attributes (location, row/column,
color, shape, relative position).
\end{verbatim}
\endgroup

\begingroup\small
\begin{verbatim}
[USER]
You are given:

* `instruction`: the task language instruction for this episode/action
                 interval
* `obs_frame`: the observation frame image (the first frame of the
               action interval)
* `contact_frame`: the contact frame image (the frame at the gripper-
                   close moment)

Task:
Generate a segmentation text query `sam3_prompt` for SAM3 that will
produce a mask of the **minimal target part** the robot is about to
manipulate **in `obs_frame`**.

Guidelines:

1. Output a **single short noun phrase** (1-10 words) suitable for
   open-vocabulary segmentation.
2. When a smaller part is clearly the interaction target, the phrase
   must refer to a **physical part** (handle/knob/lid/button/lever/
   edge/latch/hinge/tab/strap/rim) rather than a whole object.
3. Use `contact_frame` to infer the true contact target (where the
   gripper touches). Use `obs_frame` to phrase it in visible terms.
4. Preserve and include spatial qualifiers if present or inferable
   (e.g., "top-left", "second row right", "front", "leftmost",
   "upper", "nearest", "on the right side").
5. If the instruction is ambiguous, resolve it using visual evidence.
   If still ambiguous, choose the most likely minimal part and add
   one discriminative qualifier (e.g., color/position).
6. Avoid verbs and action words (open/pull/push/turn). Avoid pronouns
   ("it", "that"). Avoid long descriptions.
7. Do NOT mention "robot", "gripper", "contact", "frame", "image",
   "mask", "SAM", or "segmentation".
8. If the target is a drawer/door, prefer **handle** or **edge**. If
   it's a button/switch, prefer **button**/**switch**. If it's a lid,
   prefer **lid tab**/**lid edge**. If it's a black cup, prefer
   **black cup handle**.

Output format (strict):
Return ONLY a JSON object with two keys:
{"rationale": "<your rationale>", "sam3_prompt": "<your noun phrase>"}

Where:

* `rationale` is a very concise and very brief explanation of your
  reasoning.
* `sam3_prompt` is the best phrase.

Hard-Fail Policy (must follow):
- You output {"rationale": "<your rationale>", "sam3_prompt": null}
  ONLY when the instruction and the images are fundamentally
  incompatible such that selecting a manipulable target part would
  be guesswork.
- "Fundamentally incompatible" means: the instruction refers to an
  object/affordance category that is not present in obs_frame AND
  there is no clear gripper-contact target in contact_frame that
  could plausibly satisfy the instruction.
- Do NOT fail just because multiple candidates exist; only fail if
  it is genuinely impossible to identify any plausible target part.

Hard-Fail Affordance-Relevance Filter (must follow):
- You output {"rationale": "<your rationale>", "sam3_prompt": null}
  when the task is NOT useful for affordance model training. This
  includes:
  1. **No clear physical manipulation target**: the task does not
     involve contacting and manipulating a specific, localizable
     part (e.g., "move to the left", "wait", "look around",
     "navigate to the kitchen").
  2. **Ambiguous / unresolvable target**: even with both images, it
     is impossible to determine a single, well-defined contact
     region — e.g., the instruction is too vague ("do something with
     the stuff on the table") and the images provide no
     disambiguating evidence.
  3. **Non-rigid / deformable / soft-body object**: the target is a
     deformable, soft, or fabric-like object that lacks a well-
     defined rigid part structure. Our project focuses on rigid
     objects with strong part-level affordances (handles, knobs,
     lids, buttons, etc.). Filter out tasks involving cloth, fabric,
     towels, rope, dough, sponges, or similar soft bodies (e.g.,
     "fold the towel", "hang the cloth", "flatten the dough",
     "squeeze the sponge"). Exception: if the task involves grasping
     a rigid part OF a soft object (e.g., a zipper pull on a
     jacket), keep it.
  4. **Trivial / non-functional interaction**: the task does not
     teach a meaningful affordance mapping — e.g., the instruction
     directly names the exact contact part rather than describing a
     functional task ("grasp the handle", "touch the knob"), or the
     task is purely a sensor/state check with no physical
     manipulation.
  5. **Pick-and-place / whole-object relocation**: the task is
     simply grasping an entire object and moving it to a different
     location. These do not teach meaningful part-level affordances
     because (a) the contact region is any graspable surface rather
     than a specific functional part, and (b) the post-contact
     trajectory is generic relocation (lift -> translate -> place)
     rather than a functionally determined motion (pull, rotate,
     press, flip, slide, etc.). Filter out instructions that
     describe picking up, moving, transferring, or placing an object
     from one location to another — e.g., "move the cup to the
     left", "pick up the apple and put it in the bowl", "place the
     block on the shelf", "put the bottle on the counter",
     "transfer A to B", "stack the cubes", "sort the objects into
     the bin". Exception: keep the task if it requires interacting
     with a specific functional part to achieve the relocation
     (e.g., "pick up the pot by its handle" names a functional
     grasp point; "slide the drawer out" involves a handle/edge
     affordance).
- You strictly filter. Leave out borderline samples. If a task even
  partially matches one of the above categories, output null. High-
  quality training data is far more valuable than quantity — a noisy
  sample hurts the model more than a missing one. Only output a
  valid sam3_prompt when you are confident the task has a clear,
  rigid, localizable manipulation target with meaningful task-level
  semantics.

instruction:
{instruction}
\end{verbatim}
\endgroup

\paragraph{Curve fit.}
\label{sec:appendix-curve}
\label{sec:appendix-curvefit}
\label{sec:appendix-backproj}
For each valid object track, we fit the Bézier supervision used in
Sec.~\ref{sec:method_motion} from the recovered 3D mask-centroid
trajectory. Let $\{\mathbf{x}_i\}_{i=1}^{N}$ denote the back-projected object
positions ordered by frame. Because these points are affected by depth
noise, mask jitter, and occasional tracking jumps, we first detect
abnormally large temporal steps, down-weight them, and smooth local
neighbourhoods with a robust geometric median. The smoothed trajectory is
then resampled into a small set of approximately uniform support points
along cumulative arc length, with weights inversely proportional to the
local spatial spread. We fit a planar constant-curvature primitive to
these support points by nonlinear least squares with a Cauchy robust loss:
the primitive is parameterized by an origin $\mathbf{p}_0$, an orthonormal frame
$(\mathbf{e}_1,\mathbf{e}_2,\mathbf{n})$, curvature $\kappa$, and monotone arc-length
coordinates $s_i$, so that
$\hat{\mathbf{x}}(s)=\mathbf{p}_0+s\,\mathrm{sinc}(\kappa s)\mathbf{e}_1
+s\,\mathrm{cosc}(\kappa s)\mathbf{e}_2$. If
$|\kappa|$ times the fitted arc length is below a small threshold, the
primitive is snapped to a straight line, which avoids overfitting nearly
linear motions. The fitted curve is sampled densely and converted to the
canonical cubic Bézier target by solving a least-squares problem for the
two interior control points while fixing the start and end points. We
store the resulting control points relative to the contact anchor
$\vP_0$, matching the model output in Eq.~\ref{eq:bezier}; the raw
trajectory is preserved only for diagnostics and visualization.

\subsection{Gallery on the \ourdataset}
\label{sec:appendix-ourmodeltestset-gallery}

\begin{figure*}[p]
  \centering
  \includegraphics[width=\textwidth]{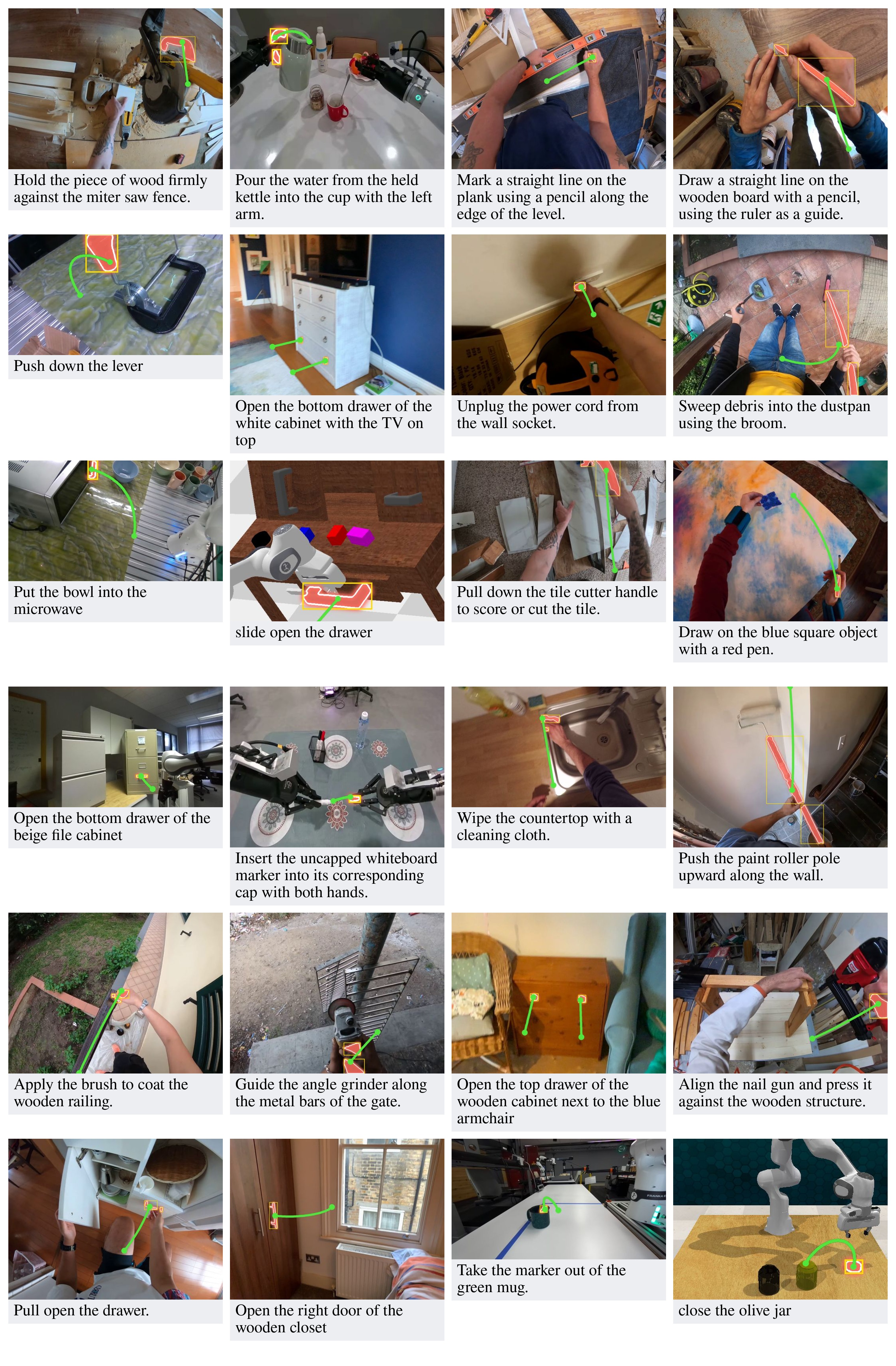}
  \caption{Qualitative gallery on
    \ourdataset, Part 1.}
  \label{fig:qual-gallery-3d-motion-a}
\end{figure*}

\begin{figure*}[p]
  \centering
  \includegraphics[width=\textwidth]{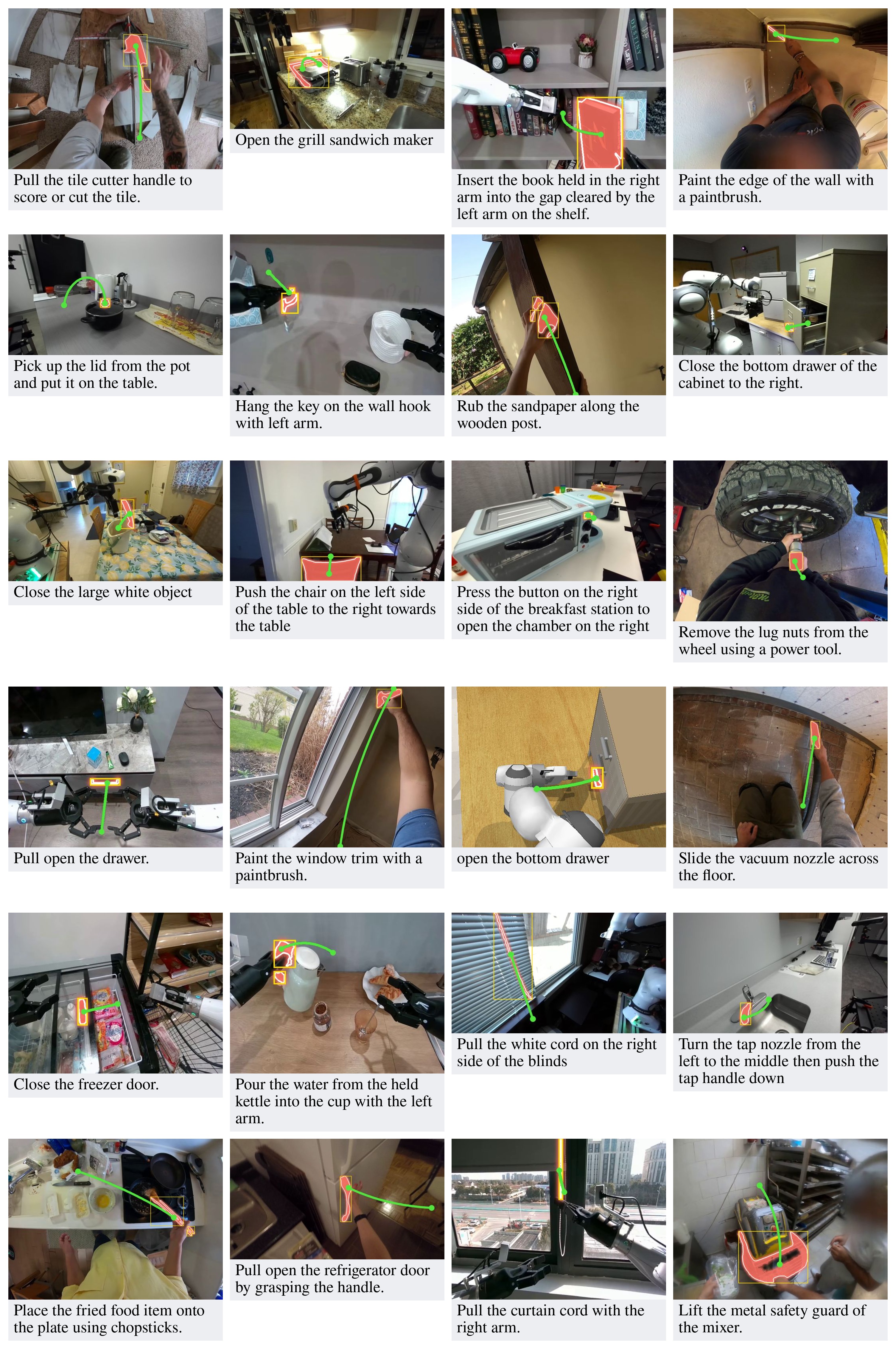}
  \caption{Qualitative gallery on 
    \ourdataset, Part 2.}
  \label{fig:qual-gallery-3d-motion-b}
\end{figure*}

We sampled 48 entries from \ourdataset and
visualize each as the observation frame with the SAM3 affordance mask
(red mask + yellow bounding box) and the \curverep
fitted 3D trajectory (green curve). Samples are split across
Fig.~\ref{fig:qual-gallery-3d-motion-a} and
Fig.~\ref{fig:qual-gallery-3d-motion-b}; the language
instruction for each sample is shown directly below its image.

\newpage
\subsection{Converting HOVA-500K for Segmentation Training}
\label{sec:appendix-hova500k-seg}

As described in \emph{Stage~2: End-to-End Training for Affordance
Segmentation} of Sec.~\ref{sec:method_training}, our affordance
segmentation model is trained on HOVA-500K~\citep{gloverpp},
RAGNet~\citep{ragnet},
InstructPart~\citep{instructpart}, and ReasonAFF~\citep{affordancer1}.
RAGNet, InstructPart, and ReasonAFF already match our required format:
an RGB image, a task query, and a binary affordance mask. HOVA-500K
instead provides point-level contact supervision. In our loader, each
HOVA sample is normalized to an image, an object noun field, a verb
field, and a contact point. The
contact point is recovered from the peak of the Gaussian contact heatmap
for 3DOI, Ego4D, and HANDAL, and from the mean of annotated contact
points for EPIC-100. This is useful affordance evidence, but it must be
converted to dense masks before training our segmentation decoder.

To turn the point annotation into a mask, we run a single-frame version
of our affordance segmentation annotation pipeline. Qwen~\citep{qwen35blog} model receives the image,
the noun and verb fields, and the contact point, producing a compact part-level prompt for SAM3, such as ``drawer handle''. SAM3 then segments the image with this prompt. We keep a
mask only if it is both confident and spatially consistent with the HOVA
contact point: among masks with confidence above $0.5$, we choose the
highest-ranked mask whose centroid lies within $0.07W$ pixels of the
contact point, where $W$ is the image width. Samples without such a mask
are rejected.

The part-level SAM3 prompt is used only to obtain the mask; it is not
used as the training query, since directly naming the contacted part
would leak the answer. We therefore run a second Qwen~\citep{qwen35blog}
 pass on the
original image and the selected-mask overlay. This pass rewrites the
sample into a natural task-level instruction that implies the affordance without naming the highlighted region explicitly. 

\section{Extended Evaluation Qualitative Results}
\label{sec:app_eval}

\begin{figure*}[p]
  \centering
  \includegraphics[width=\textwidth]{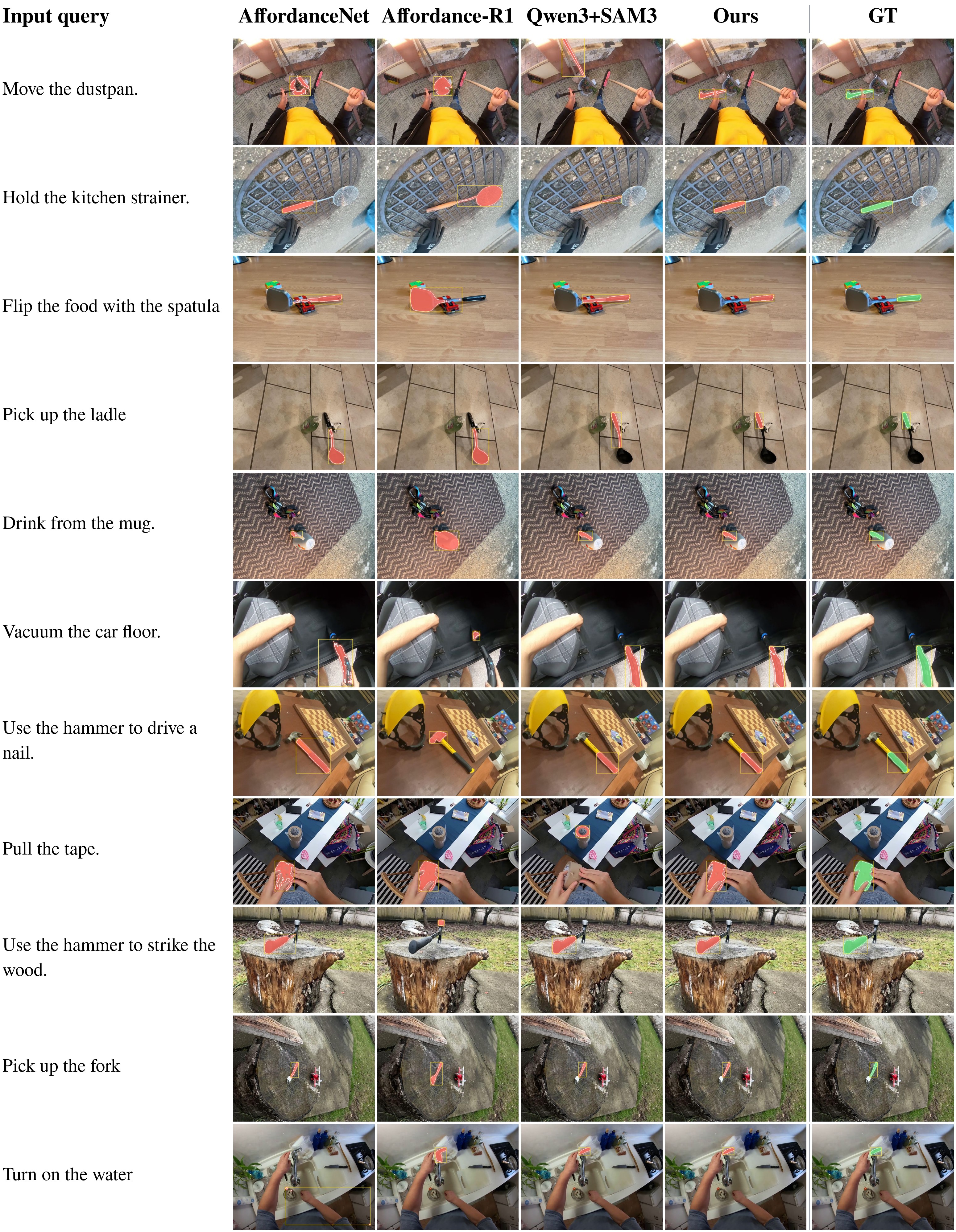}
  \caption{Qualitative gallery on the affordance segmentation evaluation, Part 1.}
  \label{fig:qual-gallery-hova}
\end{figure*}

\begin{figure*}[p]
  \centering
  \includegraphics[width=\textwidth]{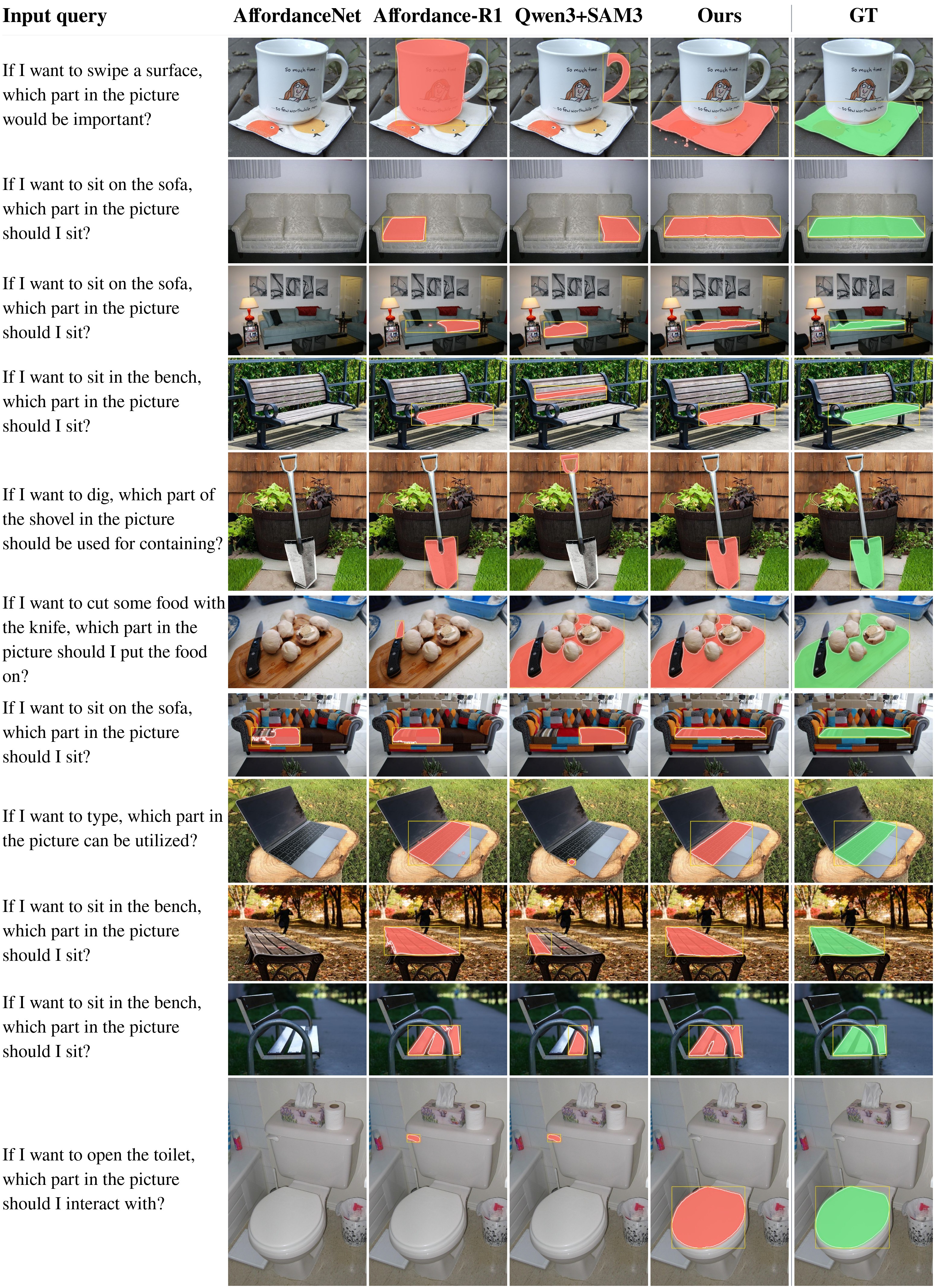}
  \caption{Qualitative gallery on the affordance segmentation evaluation, Part 2.}
  \label{fig:qual-gallery-instruct-a}
\end{figure*}

\begin{figure*}[p]
  \centering
  \includegraphics[width=\textwidth]{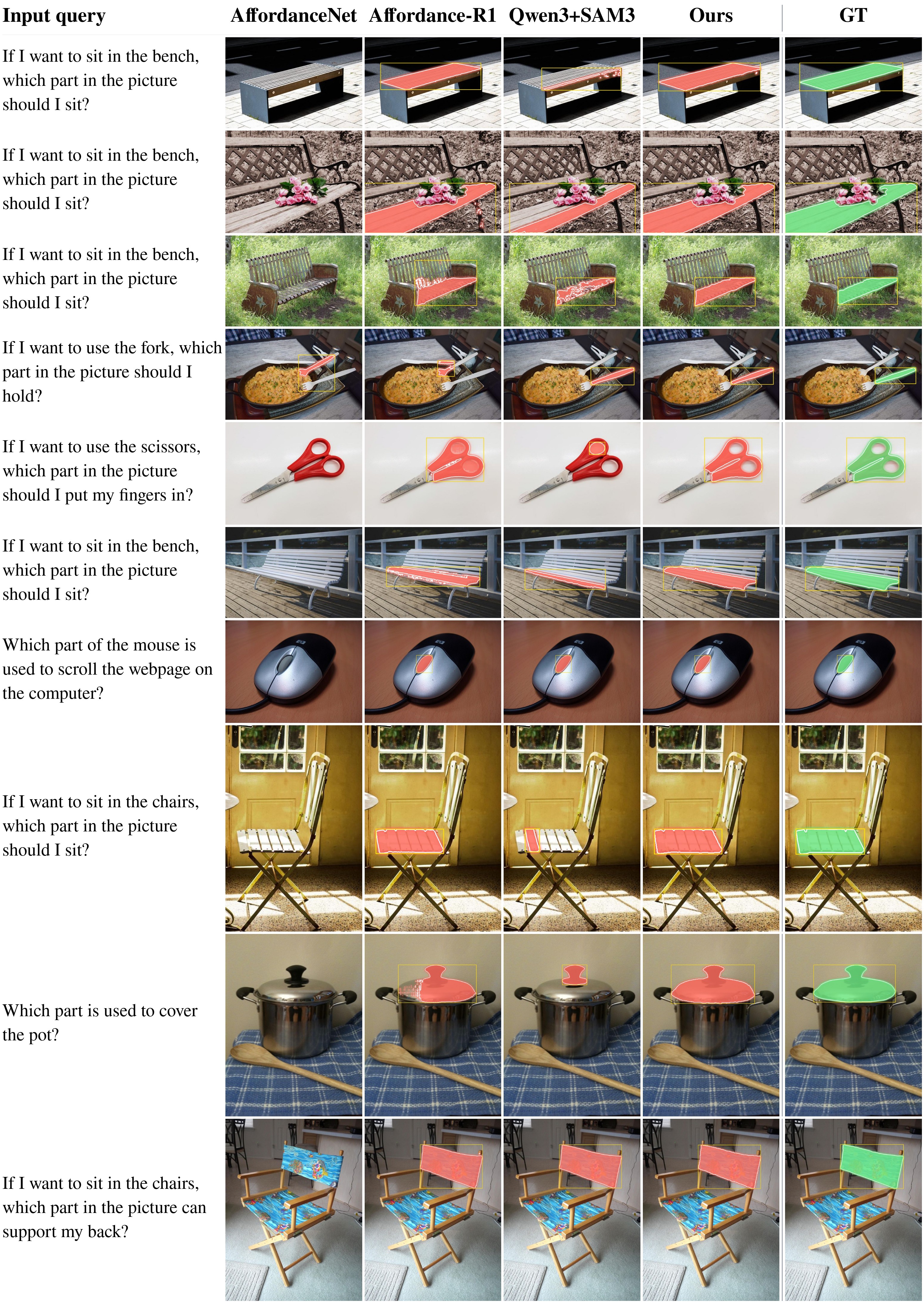}
  \caption{Qualitative gallery on the affordance segmentation evaluation, Part 3.}
  \label{fig:qual-gallery-instruct-b}
\end{figure*}

\begin{figure*}[p]
  \centering
  \includegraphics[width=\textwidth]{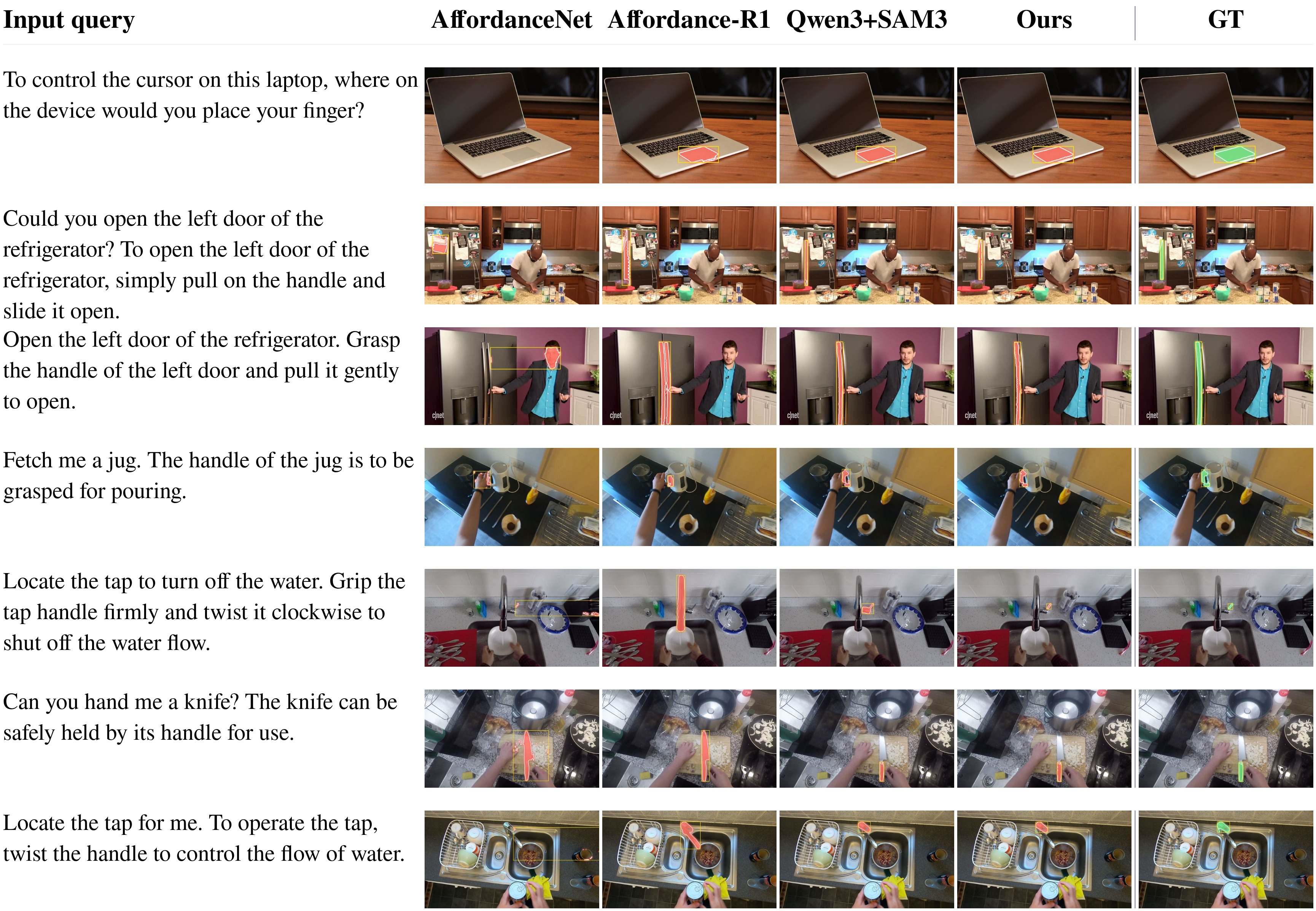}
  \caption{Qualitative gallery on the affordance segmentation evaluation, Part 4.}
  \label{fig:qual-gallery-reasoning}
\end{figure*}

\paragraph{Extended Qualitative gallery on the affordance segmentation evaluation.}
From Fig.~\ref{fig:qual-gallery-hova} to \ref{fig:qual-gallery-reasoning}, we provide qualitative galleries on the affordance segmentation evaluation results across \ourmodel, AffordanceNet, Affordance-R1, and Qwen3+SAM3. Each row shows the input query, predictions from
  three baseline models, \ourmodel{}, and the
  ground-truth mask. Red overlays indicate predicted regions, green
  overlays indicate ground truth, and yellow boxes mark mask boxes.

\begin{figure*}[p]
  \centering
  \includegraphics[width=\textwidth]{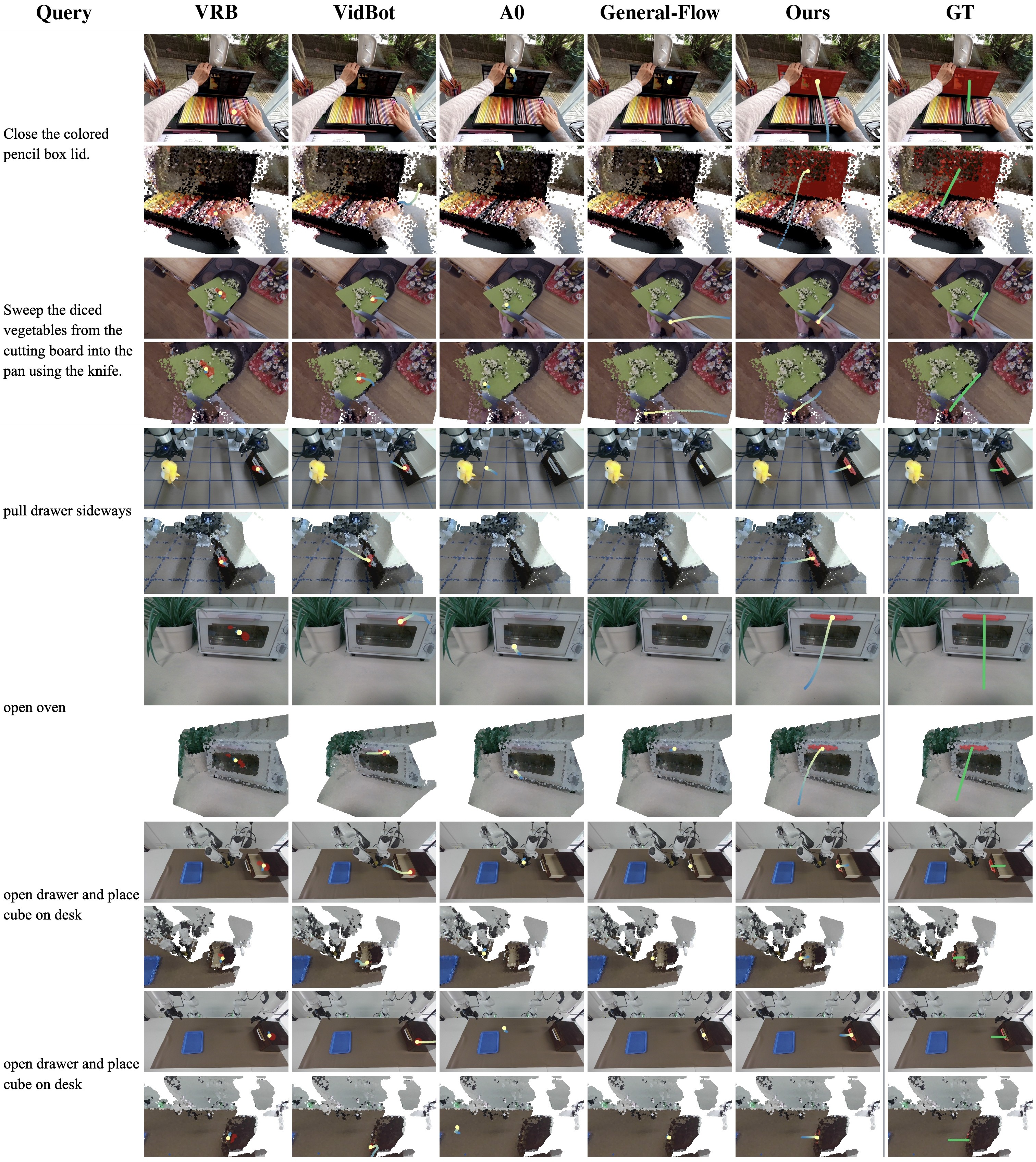}
  \caption{Qualitative gallery on the 3D motion evaluation, part 1.}
  \label{fig:qual-gallery-3d-motion-1}
\end{figure*}

\begin{figure*}[p]
  \centering
  \includegraphics[width=\textwidth]{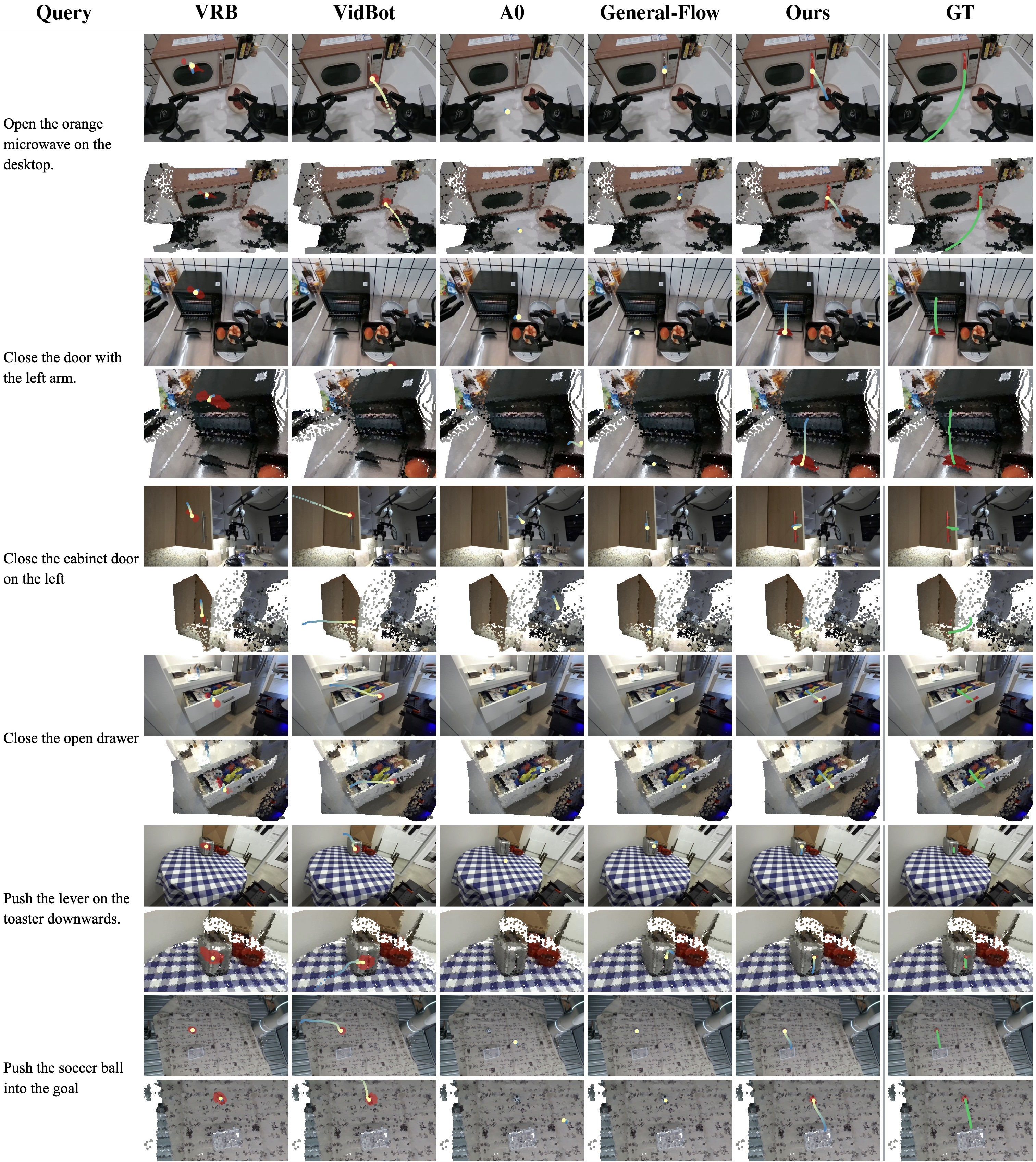}
  \caption{Qualitative gallery on the 3D motion evaluation, part 2.}
  \label{fig:qual-gallery-3d-motion-2}
\end{figure*}

\begin{figure*}[p]
  \centering
  \includegraphics[width=\textwidth]{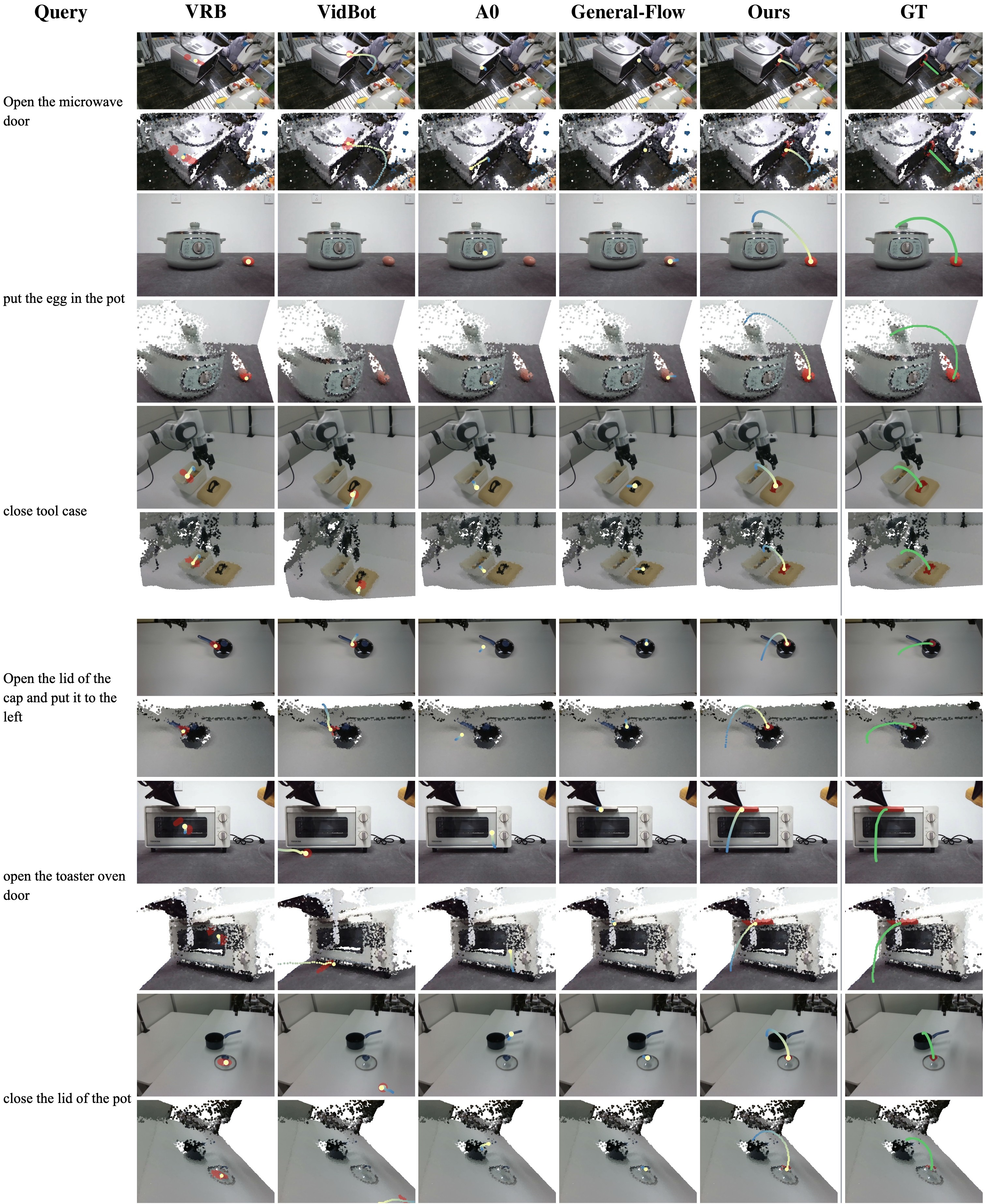}
  \caption{Qualitative gallery on the 3D motion evaluation, part 3.}
  \label{fig:qual-gallery-3d-motion-3}
\end{figure*}

\begin{figure*}[p]
  \centering
  \includegraphics[width=\textwidth]{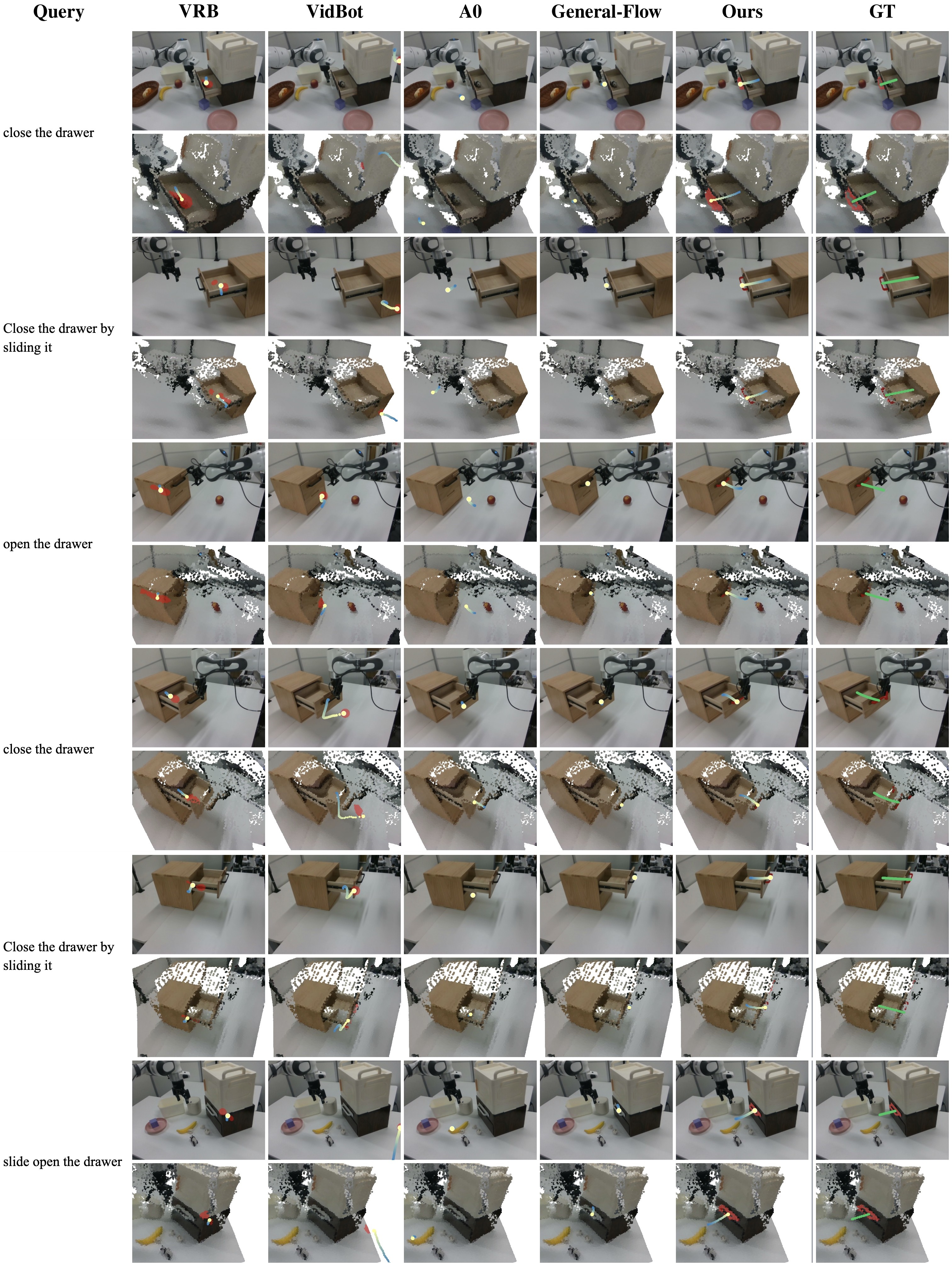}
  \caption{Qualitative gallery on the 3D motion evaluation, part 4.}
  \label{fig:qual-gallery-3d-motion-4}
\end{figure*}

\begin{figure*}[p]
  \centering
  \includegraphics[width=\textwidth]{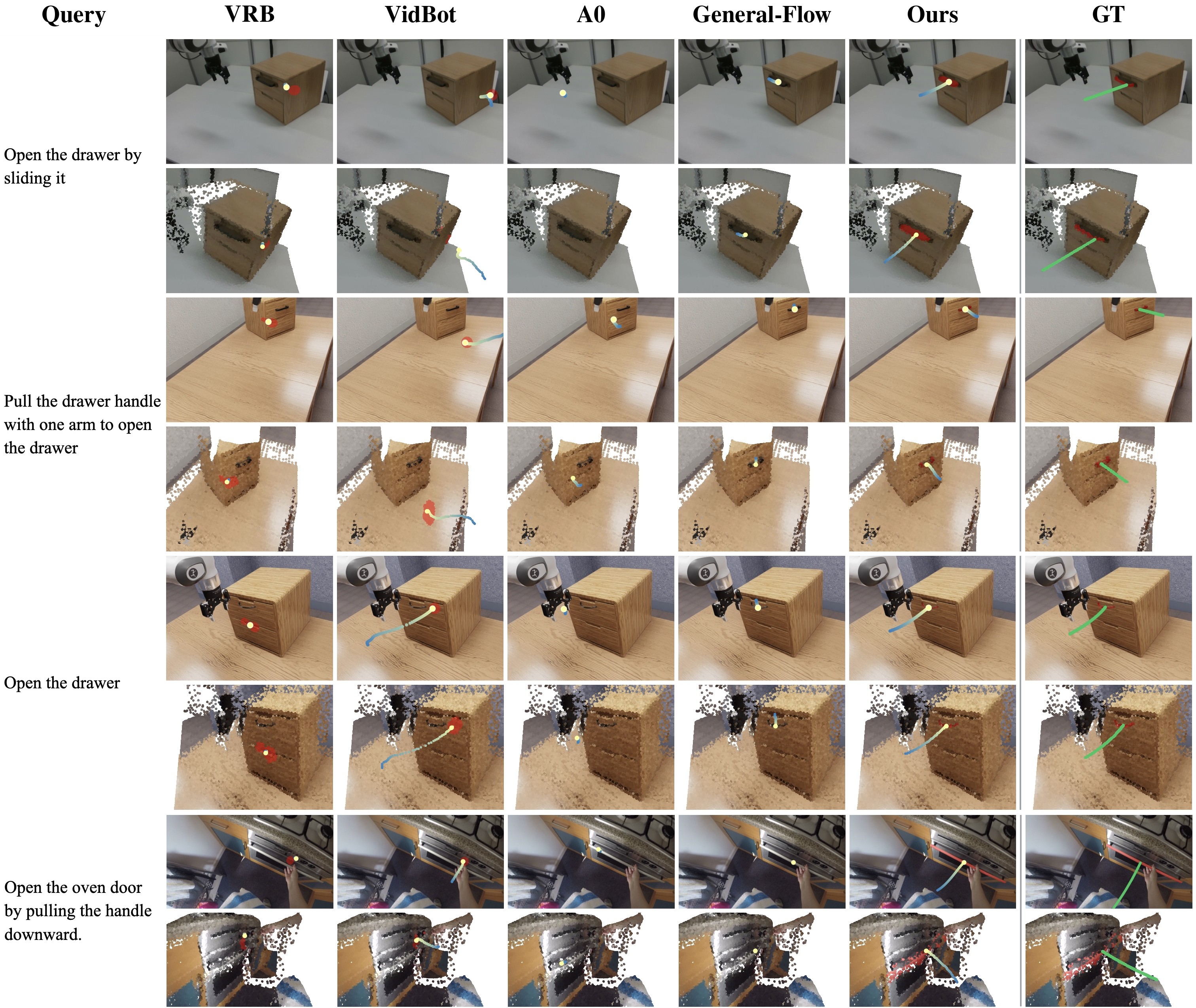}
  \caption{Qualitative gallery on the 3D motion evaluation, part 5.}
  \label{fig:qual-gallery-3d-motion-5}
\end{figure*}

\paragraph{Extended Qualitative gallery on the 3D motion evaluation.}
From Fig.~\ref{fig:qual-gallery-3d-motion-1} to \ref{fig:qual-gallery-3d-motion-5}, we additionally provide a qualitative gallery for the 3D motion evaluation across \ourmodel, VRB, VidBot, A0, and General-Flow. Each row shows the input task query in the
leftmost column, followed by predictions from four baseline models, \ourmodel{}, and the ground-truth annotation in the
rightmost column. Within each cell, the top tile is the predicted
trajectory and mask projected onto the input frame, and the bottom tile
is the back-projected 3D point cloud with the same overlays. Predicted
trajectories are coloured yellow to blue from start to end; the ground-truth trajectory is rendered as a
green curve.

\begin{figure}[!h]
    \centering
    \includegraphics[width=0.8\linewidth]{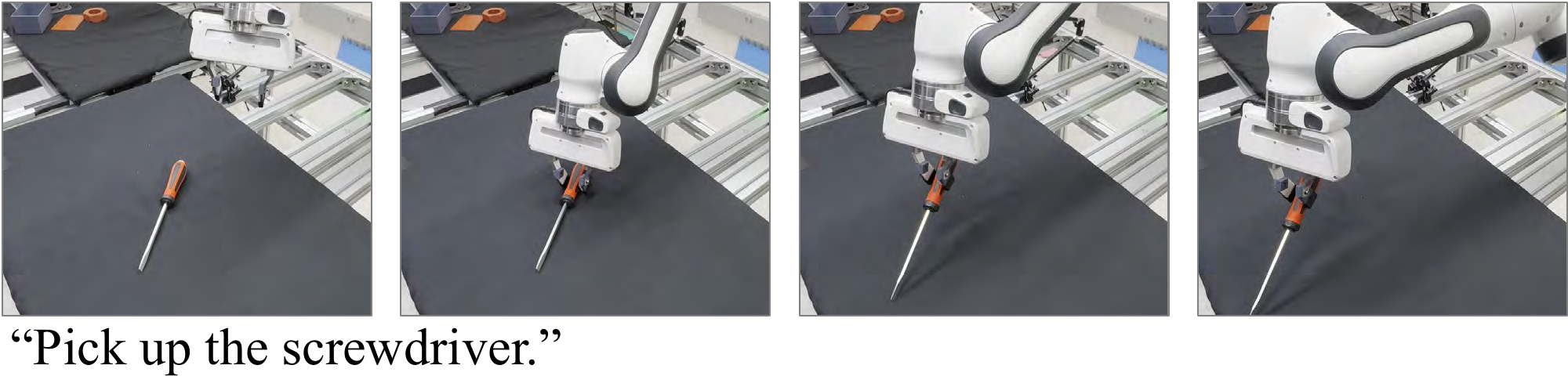}
    \caption{This example shows the grippers with our model successfully locating the action part of the screwdriver (handle) and pick it up.}
    \label{fig:additional_demo}
\end{figure}

\paragraph{Additional Qualitative results of Robotic Demo.}
We show additional results of the robotic demo, in which \ourmodel is instructed to provide trajectories to pick up the screwdriver in the scene (Figure~\ref{fig:additional_demo}).

\clearpage

\clearpage  


\end{document}